\documentclass{article}

\PassOptionsToPackage{numbers, compress}{natbib}

\usepackage[dblblindworkshop, final]{neurips_2026}

\usepackage[utf8]{inputenc}
\usepackage[T1]{fontenc}
\usepackage{times}
\usepackage{latexsym}
\usepackage{inconsolata}
\usepackage{microtype}

\usepackage{amsmath}
\usepackage{amssymb}   
\usepackage{amsthm}
\usepackage{nicefrac}

\usepackage{graphicx}
\usepackage{wrapfig}
\usepackage{float}
\usepackage{subcaption}
\usepackage{booktabs}
\usepackage{multirow}
\usepackage[table]{xcolor}

\usepackage{pifont}
\newcommand{\xmark}{\ding{55}}

\usepackage{url}
\usepackage{hyperref}
\usepackage{cleveref}

\title{Generalization Is Stability, Not Accuracy: Multi-Axis Evaluation of LLMs}
\workshoptitle{TAE (Trust-AI-Eval): Can We Trust AI Evaluation?}

\author{%
  \textbf{Nagham Omar}\thanks{Equal contribution.} \quad
  \textbf{Mahmoud Jabarin}\footnotemark[1] \quad
  \textbf{Maya Rozenshtein}\footnotemark[1] \quad
  \textbf{Rom Himelstein} \quad \\
  \textbf{Avi Mendelson} \quad
  \textbf{Amit LeVi} \\[4pt]
  Technion -- Israel Institute of Technology \\[2pt]
  \texttt{\{nagham.omar, mahmoud.j, rmaya, romh, amitlevi\}@campus.technion.ac.il} \\
  \texttt{mendlson@technion.ac.il}
}

\begin{document}

\maketitle




\begin{abstract}
\label{abs:summary}
Generalization in large language models (LLMs) is the ability to produce consistent and semantically stable outputs when the same input is expressed in different ways. Existing work typically evaluates generalization through aggregate accuracy on a single prompt format, task, or set of variations, which conflates robustness with overall benchmark performance. In this work, we show generalization evaluation at the level of individual examples, across multiple input variants, and across different aspects of model behavior, focusing on variability rather than reducing performance to a score that can be improved through narrow training or other ways that obfuscate generalization evaluation. Following this view, we introduce the Stability-Aware Generalization Objective (SAGO), a framework that measures how much model behavior changes for the same input under different variations and benchmarks, capturing variability across several dimensions including generation consistency, internal activations, confidence, and response mirroring. We show that many commonly used models exhibit statistically significant and consistent generalization instability: no model generalizes uniformly, behavioral axes capture independent failure modes, and cross-dataset variation can reverse model rankings. A reference implementation is available at \href{https://github.com/Mahmoudbj48/SAGO/tree/paper}{\includegraphics[height=1em]{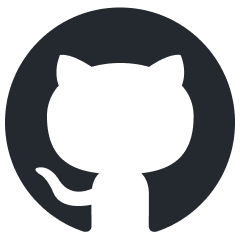}}.
\end{abstract}
\section{Introduction}
\label{sec:introduction}

LLMs are used in real-world systems, including search~\citep{hersh2024search, xiong2024search, narayanan2025search, zhu2025large}, coding assistants~\citep{ross2023programmer, coello2024effectiveness, husein2025large, zhang2026survey}, and agentic workflows~\citep{brohi2025research, plaat2025agentic}. In these settings, the same user intent is rarely expressed in one fixed form. A request can vary in wording, length, tone, formatting, or noise while preserving the same goal and context.

Accordingly, generalization is the ability of a model to preserve performance across such semantically equivalent input variations \citep{elazar2021measuring, sclar2024quantifying, zhuo2024prosa}.  This requirement stems from several factors; for instance, pretraining and post-training can expose a model to many examples, but they cannot cover all valid ways a user may express the same request~\citep{lunardi2025robustness, truong2025persona}. 
Yet, an aligned model should remain useful, solve tasks, refuse harmful requests, and follow its policies regardless of how the context is expressed in the input~\citep{ouyang2022training, wei2023jailbroken}.


Previous work studies related forms of generalization robustness, such as distribution shifts \citep{koh2021wilds, wang2023decodingtrust}, meaning-preserving perturbations \citep{elazar2021measuring, errica2025did}, and prompt sensitivity to wording, formatting, tone, input length, and input order \citep{lu2022fantastically, levy2024same, mizrahi2024multiprompt, salinas2024butterfly, sclar2024quantifying, zhuo2024prosa}, together establishing that LLM behavior shifts even when task meaning is preserved.

\begin{figure*}
\centering
\includegraphics[width=0.9\linewidth]{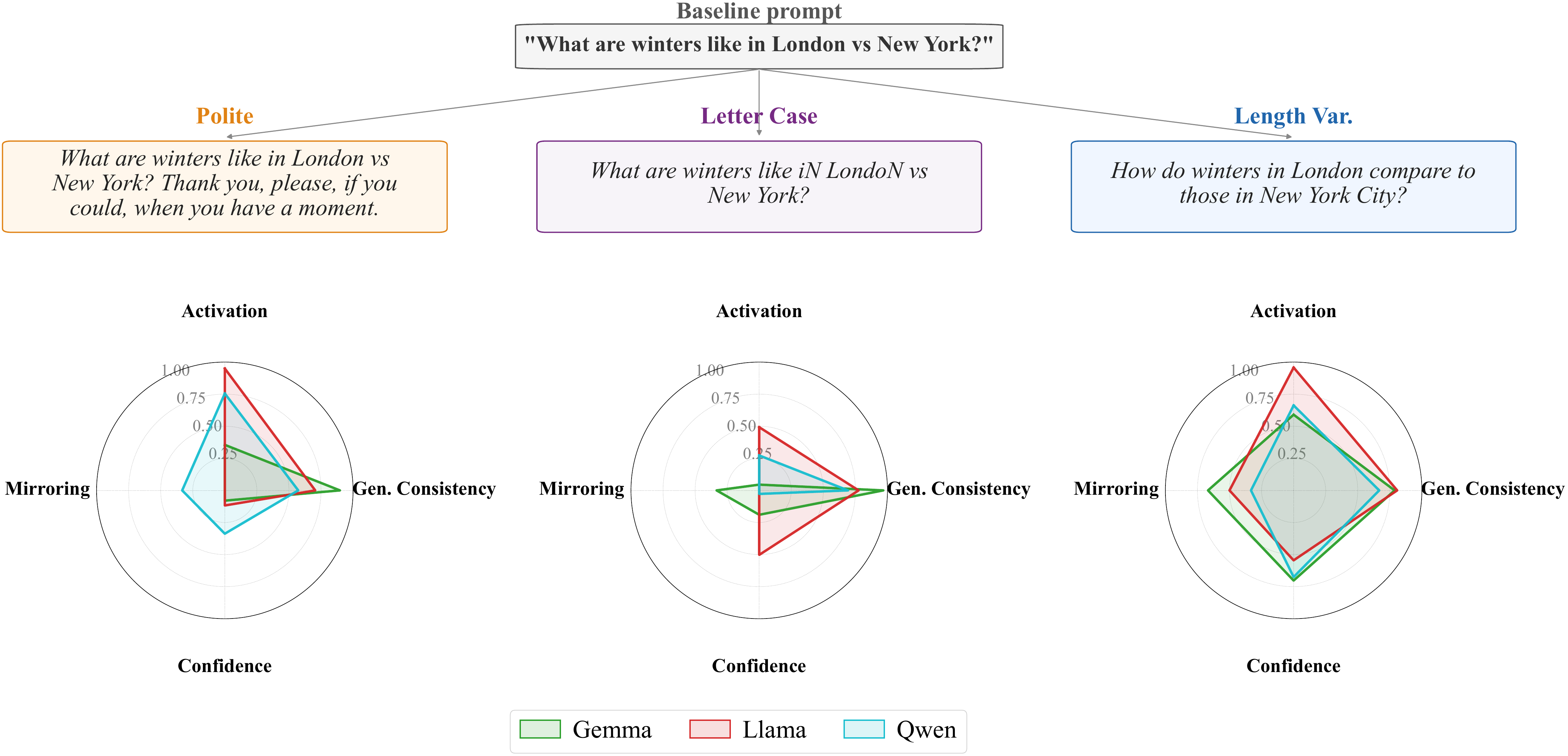}
\caption{Semantically equivalent factual prompts induce measurable behavioral shifts across model families. Radars show mean delta magnitude over four axes.}
\label{fig:motivation}
\end{figure*}

However, while these works provide important insights into robustness under specific settings and perturbations, they do not fully capture a model’s overall generalization ability for several reasons. First, many existing evaluations focus on a single domain. Although robustness in domains such as safety is important, success in one field cannot be considered evidence of generalization ability. Second, many studies evaluate only a particular family of semantically equivalent rewrites, so a model may appear robust because it has adapted to one variation pattern while still failing under other forms of expression. Lastly, evaluations are commonly based on aggregate benchmark scores, which can hide substantial behavioral differences at the level of individual examples. A model may maintain similar overall performance because failures on some semantically equivalent variants are offset by successes on others, even though its behavior changes considerably for individual inputs. Consequently, aggregate metrics can give a misleading sense of stable and reliable generalization.




In this work, we address this gap by introducing the \textbf{Stability-Aware Generalization Objective (SAGO)}, a framework for evaluating LLM generalization as \textbf{per-instance behavioral stability} rather than aggregate benchmark performance. Instead of asking only whether a model performs well on average, SAGO tests whether the model preserves its behavior when the same goal and context are expressed through semantically equivalent input variations. The framework combines multiple datasets, multiple families of input variation, and multiple behavioral axes, allowing us to capture failures that single-domain, single-perturbation, or aggregate-score evaluations can miss. For each prompt, we compare the model’s baseline behavior with its behavior under controlled variations and measure changes in generation consistency, internal activations, confidence and uncertainty, and response mirroring. This enables us to identify cases where a model appears stable at the benchmark level while changing substantially on individual examples, or where it is robust to one type of variation but unstable under another.

Using SAGO, we evaluate eleven open- and closed-source LLMs across six datasets and show that generalization instability is widespread, heterogeneous, and not captured by any single metric or domain alone. Our contributions are threefold:

\begin{itemize}
\item A framework for evaluating LLM generalization as per-instance stability across semantically equivalent input variations, multiple datasets, and multiple families of variation.

\item A multi-axis metric that measures behavioral change in activation geometry, generation consistency, confidence and uncertainty, and response mirroring. We treat mirroring as a behavioral sensitivity that may reflect either intended alignment or unintended instability.

\item An empirical study of eleven LLMs across six datasets showing that no model generalizes uniformly, that generation consistency is the most sensitive axis, and that content stability and response mirroring are independent dimensions of behavioral sensitivity.
\end{itemize}

\section{Related Work and Background}
\label{sec:related}

Prior work on LLM generalization robustness spans several lines, each addressing one aspect of the broader question. We organize them according to the dimensions along which generalization can be measured.

\paragraph{Out-of-distribution robustness.}
A first line of research measures how models behave when test data differs from training data. Domain generalization formalizes this as a single number summarizing performance on a held-out distribution~\citep{zhou2022domain}; large-scale benchmarks aggregate this gap over many domains~\citep{hendrycks2020measuring, koh2021wilds}, and broader evaluations of trustworthiness report aggregate accuracy under such shifts~\citep{wang2023decodingtrust}. These studies establish that distribution shifts degrade performance and that the degradation is meaningful at the dataset level. However, they do not ask whether a model produces consistent behavior on the same input expressed in different ways, and they aggregate over examples; thus, per-instance flipping is not measurable from their reported metrics.

\paragraph{Consistency and sensitivity under prompt-level variation.}
A second line of work tests whether predictions remain stable when an input is rephrased without changing its meaning. This includes identifying inconsistencies in paraphrased factual queries~\citep{elazar2021measuring} and systematic disagreement in instruction-tuned settings~\citep{salinas2024butterfly, errica2025did}. A large body of work narrows this question to specific semantic scopes, such as cross-lingual variants~\citep{deng2024multilingual, wang2024all, pattnayak2026indicjr}, social-tone shifts~\citep{yin2024should,cai2025tone, dobariya2025tone}, and consistency for risky or harmful requests~\citep{chao2024jailbreakbench, chen2024rmcbench, han2024medsafetybench, mazeika2024harmbench, wang2025call}. Parallel to these semantic rewrites, a third line of work investigates sensitivity to surface-level choices that have no intended impact on intent, such as formatting and template selection~\citep{sclar2024quantifying, zhuo2024prosa}, in-context example ordering~\citep{lu2022fantastically}, and low-level perturbations like punctuation and whitespace~\citep{abedin2025arithmattack, himelstein2025silent, seleznyov2025punctuation}. While these studies collectively establish that LLM behavior is rarely invariant under semantically equivalent inputs, they do not provide a complete evaluation of a model’s overall generalization ability. Most works focus on a single domain or a single family of input variation in isolation, making it unclear whether robustness extends beyond the specific setting being evaluated, and overestimating generalization. In addition, they typically rely on aggregate performance metrics—such as mean accuracy or performance spread—to summarize behavior. This can obscure substantial per-instance instability: a model may fail on some semantically equivalent variants while succeeding on others, preserving similar overall benchmark performance despite large behavioral changes at the individual-example level.


\paragraph{Behavioral signals beyond accuracy.}
The signals we use as separate axes have each been developed in their own literatures. Hidden-state geometry has been used to track how internal representations shift with input changes~\citep{himelstein2025silent}. Semantic similarity measures such as BERTScore quantify meaning preservation between generations~\citep{zhang2019bertscore}, complementing lexical-overlap measures such as BLEU~\citep{papineni2002bleu}. Token-level log-probabilities and predictive entropy are standard tools for measuring confidence and uncertainty~\citep{farquhar2024detecting, ma2025estimating}. Stylistic accommodation—the degree to which model output mirrors input register—has been studied as a behavioral phenomenon in its own right~\citep{truong2025persona, blevins2026language}. While mirroring is often an intended feature of instruction tuning to improve human-computer interaction, it represents a core axis of behavioral sensitivity that can also manifest as unintended instability when models mirror non-semantic variations like input noise. These signals have been used individually in service of single-axis claims; combining them within a single per-instance analysis has not been attempted.

While exceptional large-scale benchmarks successfully evaluate across multiple datasets, variation families, and non-aggregate metrics~\citep{habba2025dove}, they remain restricted to a single behavioral signal such as accuracy. This makes it difficult to determine whether instability is a property of the model more broadly, or whether it is localized to particular datasets, variation families, behavioral axes, or individual examples. A model can therefore appear stable under one evaluation view while failing under another: it may preserve average performance while changing responses for specific prompts, remain robust to one type of variation while failing under another, or show stable outputs while its confidence or internal representations shift. Thus, the missing piece is not another single robustness test, but a per-instance, multi-axis evaluation that measures how different forms of behavioral instability co-occur across datasets and semantically equivalent input variations.

\section{Stability-Aware Generalization Objective (SAGO)}
\label{sec:method}

SAGO evaluates robustness to semantically equivalent input variations: a model generalizes only if its behavior remains stable across meaning-preserving variants of the same prompt. We measure this across variation families, content domains, and individual prompts by comparing behavior on original and modified inputs along multiple behavioral axes. The framework includes: (1) a benchmark of controlled semantic variations, (2) a per-prompt stability score, and (3) behavioral axes capturing distinct failure modes.  Section~\ref{subsec:problem} formalizes the objective, Section~\ref{subsec:families} defines the variation families, Section~\ref{subsec:sgs} introduces the score and statistical
test, and Section~\ref{subsec:metrics} instantiates the four behavioral axes.


\subsection{Notation and Objective}
\label{subsec:problem}

\paragraph{Notation.}
Let $\mathcal{X}$ be the space of prompts, $\mathcal{Y}$ the
space of model responses, and
$f_\theta:\mathcal{X}\rightarrow\mathcal{Y}$ an
instruction-tuned language model evaluated over a collection of
datasets $\{\mathcal{D}_j\}_{j=1}^{J}$. For a prompt
$x_i \in \mathcal{D}_j$, the baseline response is
$y_i = f_\theta(x_i)$. Let
$c:\mathcal{X}\rightarrow\mathcal{C}$ map each prompt to its
semantic content. For each prompt we construct a shared set of
semantically equivalent variations indexed by
$v \in \mathcal{V}$:
\begin{equation}
    x_i^{(v)} = T_v(x_i), \qquad
    c\!\left(x_i^{(v)}\right) = c(x_i),
    \label{eq:variant}
\end{equation}
where $T_v:\mathcal{X}\rightarrow\mathcal{X}$ is a
meaning-preserving transformation. The construction of
$\mathcal{V}$ is described in Section~\ref{subsec:families}.

An \emph{axis} is a scalar property
$X(f_\theta, x) \in \mathbb{R}$ of the model on a prompt, and the
per-prompt change induced by variation $v$ is
$\Delta_X(i, v) = X(f_\theta, x_i^{(v)}) - X(f_\theta, x_i)$.
Deltas are signed: the sign preserves directional information
(e.g., negative $\Delta_{X_{\mathrm{Act}}}$ indicates
representational drift), while the RMS aggregation in
Section~\ref{subsec:sgs} captures magnitude via squaring.

\paragraph{Objective.}
Given a model $f_\theta$, datasets
$\{\mathcal{D}_j\}_{j=1}^{J}$, a variation set $\mathcal{V}$,
and an axis $X$, the SAGO objective is to determine whether
\begin{equation}
    \Delta_X(i, v) = 0 \quad
    \text{for all } i \text{ and all } v \in \mathcal{V},
    \label{eq:objective}
\end{equation}
and, when this does not hold, to quantify how far the model is
from satisfying it. Any violation is evidence of generalization
failure on $X$, and the magnitude of the violation characterizes
its severity. Because generalization on one axis does not imply
generalization on another, this condition is evaluated separately
for each axis (Section~\ref{subsec:metrics}).

\subsection{Benchmark Construction}
\label{subsec:families}

The benchmark combines multiple datasets with a shared set of
controlled input variations. Each variation is a tuple
$v = (F, u, s, p)$, where $F \in \mathcal{F}$ is a variation
family, $u \in \mathcal{U}_F$ is a subtype within that family,
$s \in \mathcal{S}_F$ its strength, and
$p \in \mathcal{P} = \{\textsc{prefix}, \textsc{suffix},
\textsc{global}\}$ its position when applicable. The full
variation set is
{\small
\begin{equation}
    \mathcal{V} = \bigl\{(F, u, s, p) \mid
    F\in\mathcal{F},\;
    u\in\mathcal{U}_F,\;
    s\in\mathcal{S}_F,\;
    p\in\mathcal{P}\bigr\}.
    \label{eq:variation_set}
\end{equation}
}
We use three families, each chosen so that stability under one
does not imply stability under another. All transformations
preserve meaning, i.e.,
$c(x_i^{(v)}) = c(x_i)$ as defined in Eq.~\ref{eq:variant}.

\paragraph{Social register.}
Changes the interpersonal framing of the prompt (politeness,
directness, greetings, hedging, gratitude markers, and negative
social framing) while preserving
meaning~\citep{yin2024should, cai2025tone, dobariya2025tone}.

\paragraph{Surface noise.}
Introduces token- and character-level changes such as irregular
spacing, punctuation noise, and capitalization shifts. Small
surface changes are known to affect model output even when task
meaning is unchanged~\citep{sclar2024quantifying,
seleznyov2025punctuation}.

\paragraph{Structural rewriting.}
Reorganizes the prompt while preserving the request, including
controlled compression or expansion of length and conversion
between interrogative and imperative
forms~\citep{levy2024same, mizrahi2024multiprompt, salinas2024butterfly, errica2025did}.

\subsection{Stability Generalization Score}
\label{subsec:sgs}

For a given axis $X$, the SGS quantifies how far the model is
from satisfying Eq.~\ref{eq:objective}. Because different axes
have different units and ranges, raw deltas are first normalized
to a common scale.

\paragraph{Normalization.}
To enable cross-axis comparison while preserving the zero
reference point required by Eq.~\ref{eq:objective}, each delta
is divided by a fixed per-axis scale factor,
$\widetilde{\Delta}_X(i, v) = \Delta_X(i, v) / R_X$.
For axes with bounded range, $R_X$ equals the theoretical
maximum of $|\Delta_X|$; for sequence log-probability, which
is unbounded, $R_X$ is set to a fixed constant derived from
the empirical range in our evaluation. This yields
$\widetilde{\Delta}_X \in [-1, 1]$ with zero corresponding
exactly to no behavioral change, and scores that are
comparable across models, axes, and studies. The specific
$R_X$ values are given alongside each metric definition in
Section~\ref{subsec:metrics}.

\paragraph{Per-prompt instability.}
For prompt $i$, instability on axis $X$ is the root mean square
(RMS) of the normalized deltas across the variation set,
$s_X(i) = \bigl(\frac{1}{|\mathcal{V}|}
\sum_{v \in \mathcal{V}} \widetilde{\Delta}_X(i, v)^{2}
\bigr)^{1/2}$.
A prompt satisfying Eq.~\ref{eq:objective} has $s_X(i) = 0$;
larger values indicate that semantically equivalent variations
produce different values of $X$. The RMS is zero only when every delta is zero, so a model that
shifts by the same non-zero amount on every variation is
correctly identified as unstable.


\paragraph{Dataset and overall score.}
For dataset $\mathcal{D}_j$, the score on axis $X$ is the average
per-prompt instability,
$\mathrm{SGS}_X^{(j)} = \frac{1}{|\mathcal{D}_j|}
\sum_{i \in \mathcal{D}_j} s_X(i)$,
and the overall score pools all datasets with equal per-prompt
weight,
$\mathrm{SGS}_X = \frac{1}{N}\sum_{j=1}^{J}
|\mathcal{D}_j| \cdot \mathrm{SGS}_X^{(j)}$,
where $N = \sum_{j=1}^{J} |\mathcal{D}_j|$.
Lower $\mathrm{SGS}_X$ indicates stronger generalization on
axis $X$; $\mathrm{SGS}_X = 0$ holds iff the model satisfies
Eq.~\ref{eq:objective} on $X$ for every prompt.

\paragraph{Statistical significance.}
To test whether a model exhibits practically meaningful
generalization failure on axis $X$, we test against a non-zero
tolerance threshold,
$H_0\!: \mathbb{E}[s_X(i)] \leq \varepsilon$ versus
$H_1\!: \mathbb{E}[s_X(i)] > \varepsilon$.
The threshold $\varepsilon > 0$ represents the fraction of the
normalized metric range below which instability is considered
negligible for practical deployment. We report results at
$\varepsilon \in \{0.01, 0.05, 0.10\}$, corresponding to 1\%,
5\%, and 10\% of the range; primary results use
$\varepsilon = 0.05$. We test with a one-sided $t$-test,
$t = (\bar{s}_X - \varepsilon) / \mathrm{SE}(s_X)$,
where $\bar{s}_X$ is the sample mean and $\mathrm{SE}(s_X)$ the
standard error of $\{s_X(i)\}_{i=1}^{N}$. Rejection of $H_0$ establishes that the model's instability
exceeds a practically meaningful threshold. Robustness across
all three $\varepsilon$ values is reported in
Appendix~\ref{app:epsilon}.

\subsection{Metrics}
\label{subsec:metrics}

$\mathrm{SGS}_X$ is computed independently for each axis $X$.
We instantiate four axes targeting distinct ways a model can
fail (Eq.~\ref{eq:objective}), each capturing a behavioral
sensitivity that no single task-accuracy metric can
distinguish~\citep{de2025metamorphic}.

\paragraph{Activation geometry ($X_{\mathrm{Act}}$).}
Internal representations may diverge across layers or prompt variants while final outputs agree \citep{xie2024calibrating}, making the model internally unstable but externally stable. Let $h_\ell(x)$ denote the last non-padding
hidden state at layer $\ell$ of an $L$-layer
model~\citep{himelstein2025silent}, and set
$\Delta_{X_{\mathrm{Act}}}(i,v) = \frac{1}{L}\sum_{\ell=1}^{L}
\cos\!\bigl(h_\ell(x_i^{(v)}),\, h_\ell(x_i)\bigr) - 1$,
where negative values indicate representational drift
($R_{\mathrm{Act}} = 2$, since cosine similarity ranges over
$[-1,1]$). We use the last hidden state because it aggregates the full input via causal masking~\citep{vaswani2017attention} and directly conditions generation in instruction-tuned settings~\citep{duan2024exploring}.

\paragraph{Generation consistency ($X_{\mathrm{BLEU}}$,
$X_{\mathrm{BERT}}$).}
A model can preserve average accuracy while producing different
responses to equivalent prompts~\citep{hua2025flaw}. We measure
similarity between the variant response
$y_i^{(v)} = f_\theta(x_i^{(v)})$ and the baseline
$y_i = f_\theta(x_i)$~\cite{papineni2002bleu,
zhang2019bertscore}, as
$\Delta_{X_{\mathrm{BLEU}}}(i,v) =
\mathrm{BLEU}(y_i^{(v)}, y_i) - \mathrm{BLEU}(y_i, y_i)$
and
$\Delta_{X_{\mathrm{BERT}}}(i,v) =
B_{F1}(y_i^{(v)}, y_i) - B_{F1}(y_i, y_i)$,
where $B_{F1}(\cdot,\cdot)$ denotes $\mathrm{BERTScore}_{F1}$,
BLEU is scored on $[0, 100]$, and BERTScore on $[0, 1]$.
Scale factors are set to cover the full metric range:
$R_{\mathrm{BLEU}} = 100$ and $R_{\mathrm{BERT}} = 1$.
Together these distinguish surface-level lexical change from
meaning-level drift, a distinction accuracy alone collapses.

\paragraph{Confidence and uncertainty ($X_{\mathrm{Prob}}$,
$X_{\mathrm{Ent}}$).}
A model can produce stable responses while becoming less
certain about them, or vice versa~\citep{zhuo2024prosa}. Let
$\log p_\theta(y \mid x)$ denote the sequence log-probability
and $H_\theta(y \mid x)$ the mean token-level
entropy~\citep{farquhar2024detecting, ma2025estimating}, and set
$\Delta_{X_{\mathrm{Prob}}}(i,v) =
\log p_\theta(y_i^{(v)} \mid x_i^{(v)}) -
\log p_\theta(y_i \mid x_i)$
and
$\Delta_{X_{\mathrm{Ent}}}(i,v) =
H_\theta(y_i^{(v)} \mid x_i^{(v)}) - H_\theta(y_i \mid x_i)$.
Token-level entropy serves as the length-invariant confidence
signal ($R_{\mathrm{Ent}} = 10$, based on the empirical range
observed in our evaluation). Sequence log-probability is
unbounded and unnormalized by response length, so longer
responses may register lower values purely from accumulating
more tokens~\citep{azaria2023internal}; we set $R_{\mathrm{Prob}} = 250$,
similarly based on the empirical range. We retain both to
compare length-sensitive and length-invariant views of
calibration drift.

\paragraph{Response mirroring ($X_{\mathrm{MR}}$).}
A model can preserve task content while adapting its response
to surface properties of the input, e.g., becoming more verbose
for longer prompts or noisier for noisy ones. Some stylistic
accommodation is an intended alignment feature (e.g., tone
matching), while other forms reflect unintended instability
(e.g., noise reproduction)~\citep{blevins2026language}.
SAGO measures mirroring as a behavioral sensitivity without
presupposing intent. We define a binary detector
$m(y, v) \in \{0, 1\}$ indicating whether response $y$
reflects the targeted surface property of variation $v$, and
set $\Delta_{X_{\mathrm{MR}}}(i,v) = m(y_i^{(v)}, v) - m(y_i, v)$, yielding values in $\{-1, 0, 1\}$
($R_{\mathrm{MR}} = 1$). The detector is family-specific
(Section~\ref{sec:exp-setting}).

\bigskip
\noindent
Together, the four axes span the path from internal
representation to surface output, capturing behavioral
sensitivities that aggregate metrics
hide~\citep{habba2025dove}: instability on
individual examples, instability under specific variation
families, and instability on one axis but not another.
\section{Experiments}
\label{sec:experiments}

We evaluate instruction-tuned LLMs under the SAGO framework
(Section~\ref{sec:method}), computing $\mathrm{SGS}_X$ for
each behavioral axis across all combinations of variation
family, subtype, strength, and position. Our evaluation
addresses three questions: whether failure modes dissociate
across the six $\Delta$-axes; whether within-model cross-dataset
variation is large enough to reverse model rankings; and whether
closed-source models exhibit a qualitatively distinct instability
profile relative to open-source alternatives.




\subsection{Experimental Setting}
\label{sec:exp-setting}



\paragraph{Setup.}
We evaluate eight open-source instruction-tuned models spanning
three families: Llama-3.2-3B-Instruct and Llama-3.1-8B-Instruct
(L-3B, L-8B)~\cite{grattafiori2024llama}; Gemma-2B-IT,
Gemma-7B-IT, and Gemma-4-E4B-IT
(G-2B, G-7B, G4-E4B)~\cite{team2024gemma}; and
Qwen2.5-1.5B-Instruct, Qwen2.5-7B-Instruct, and Qwen3.5-9B
(Q-1.5B, Q-7B, Q3.5-9B)~\cite{qwen2025qwen25technicalreport}. We
additionally evaluate three closed-source models via API:
GPT-5.4~\cite{openai2026gpt54},
Gemini-2.5-Flash (Gem-2.5F)~\cite{comanici2025gemini}, and
Claude-Sonnet-4.6 (Cl-S-4.6)~\cite{anthropic2026claude46sonnet}. All
models use deterministic decoding ($T{=}0$); $X_\mathrm{Act}$
and $X_\mathrm{Prob}$ are unavailable for closed-source models
due to API limitations, while $X_\mathrm{Ent}$ is computed
only for GPT-5.4. Experiments run on four NVIDIA RTX~2080~Ti
and four GTX~1080~Ti GPUs (CUDA~12.8); closed-source evaluation
uses provider APIs.

We evaluate on six open-ended factual QA datasets spanning
health, law, finance, politics, science, history, geography,
general knowledge, and creative instruction-following:
TruthfulQA~\citep{lin2022truthfulqa},
Natural Questions~\citep{kwiatkowski2019natural},
Alpaca~\citep{taori2023stanford-alpaca},
SimpleQA Verified~\citep{haas2025simpleqa},
TriviaQA~\citep{joshi2017triviaqa}, and
HotpotQA~\citep{yang2018hotpotqa}.
All elicit paragraph-length free-form responses suitable for
measurement on all four axes, ensuring that $\mathrm{SGS}_X$
is not estimated on a narrow content distribution. We sample $N{=}16$ prompts per dataset, justified
by the sample size ablation in
Appendix~\ref{subsec:ablation}.

\paragraph{Variation set $\mathcal{V}$.}
The variation set (Eq.~\ref{eq:variation_set}) instantiates
three families. \textit{Social register} varies prompt framing
across eleven politeness levels. \textit{Surface noise}
applies token- and character-level perturbations via three
subtypes: spacing, punctuation, and letter casing, each at
multiple strengths. \textit{Structural rewriting} reorganizes
the prompt via length variation and sentence-form conversion
(interrogative/imperative). Strength levels and examples for
all families are in Appendix~\ref{app:variation_examples}.
Non-structural variants are applied at three positions
$p \in \{\textsc{prefix}, \textsc{suffix}, \textsc{global}\}$;
structural variants are global by construction. Semantic
preservation is guaranteed by construction for surface-noise
variants; for the remaining families, we retain only prompts
where all variants exceed $\mathrm{BERTScore} > 0.85$~%
\citep{zhang2019bertscore, morris2020second, berger2021don}.
Strength and placement ablations are in
Appendix~\ref{app:sensitivity}.

\paragraph{Metric implementation.}
For $X_\mathrm{BLEU}$ and $X_\mathrm{BERT}$, BLEU is rescaled
to $[0,1]$ and BERTScore-F1 uses the RoBERTa-large backbone.
$X_\mathrm{Prob}$ and $X_\mathrm{Ent}$ are computed under
teacher forcing. For $X_\mathrm{MR}$: social-register
mirroring uses GPT-4o-mini~\citep{openai2024gpt4omini}
as judge~\citep{zheng2023judging} (10\% of labels manually
verified); surface-noise mirroring uses rule-based heuristics;
length mirroring checks whether response length matches the
target multiplier within $\pm 0.25$. Primary results use
$\varepsilon{=}0.05$; robustness across
$\varepsilon \in \{0.01, 0.05, 0.10\}$ is in
Appendix~\ref{app:epsilon}. Tables denote significance with
$^{*}p{<}0.05$, $^{**}p{<}0.01$, $^{\dagger}p{<}0.001$.

\subsection{Experimental Results}
\label{sec:exp-results}

\begin{figure}[t]
    \centering
    \begin{subfigure}[b]{0.48\textwidth}
        \centering
        \includegraphics[width=\linewidth]{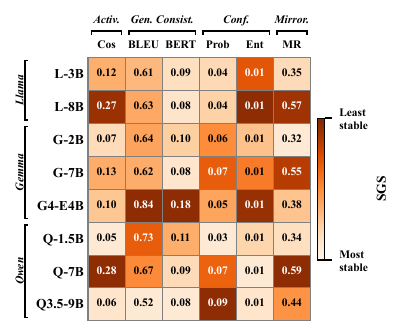}
        \caption{Total SGS, cell shading is normalised independently per column. For all open-source models across the $\Delta$-axes, pooled across all datasets.}
        \label{fig:sgs_heatmap_main}
    \end{subfigure}
    \hfill 
    \begin{subfigure}[b]{0.45\textwidth}
        \centering
        \includegraphics[width=\linewidth]{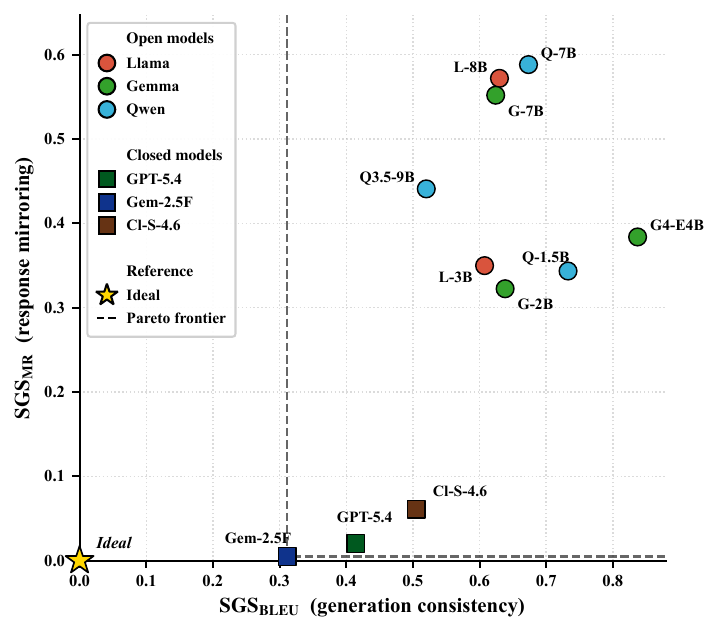}
        \caption{$\mathrm{SGS}_{\mathrm{BLEU}}$ vs.\ $\mathrm{SGS}_{\mathrm{MR}}$. For the eleven evaluated models, pooled across all datasets.}
        \label{fig:pareto_bleu_vs_mr}
    \end{subfigure}
    
    \caption{Model evaluation metrics pooled across all six datasets.}
    \label{fig:combined_models_metrics}
\end{figure}

Figure~\ref{fig:sgs_heatmap_main} and
Table~\ref{tab:sgs_closed_per_dataset} present total and
per-dataset $\mathrm{SGS}_X$ for the open-source and
closed-source models respectively;
Figure~\ref{fig:pareto_bleu_vs_mr} compares all eleven models
on generation consistency and mirroring jointly. Full
per-dataset results are in Appendix~\ref{app:full_results}.
Exceptions to $p{<}0.001$ are marked with superscripts.

\begin{wraptable}{r}{0.48\textwidth}
 \vspace{-30pt}
 \caption{Per-dataset SGS for the three closed-source models at $\varepsilon=0.05$. $^{*}p<0.05$, $^{**}p<0.01$, $^{\dagger}p<0.001$.}
    \label{tab:sgs_closed_per_dataset}
    \centering
    \tiny
    \setlength{\tabcolsep}{3pt}
    \renewcommand{\arraystretch}{1.1}
    \resizebox{\linewidth}{!}{%
    \begin{tabular}{ll ccc}
        \toprule
        \textbf{Model} & \textbf{Dataset} & \multicolumn{2}{c}{\textbf{Generation Consistency}} & \textbf{Mirroring} \\
        \cmidrule(lr){3-4} \cmidrule(lr){5-5}
         &  & $\Delta$-BLEU & $\Delta$-BERT & $\Delta$-MR \\
        \midrule
        \multirow{6}{*}{\textbf{GPT-5.4}}
        & TruthfulQA & 0.384{\tiny$\,\pm\,$0.11}$^{\dagger}$ & 0.057{\tiny$\,\pm\,$0.02} & 0.014{\tiny$\,\pm\,$0.05} \\
        & Natural Questions & 0.354{\tiny$\,\pm\,$0.12}$^{\dagger}$ & 0.049{\tiny$\,\pm\,$0.02} & 0.027{\tiny$\,\pm\,$0.07} \\
        & Alpaca & 0.467{\tiny$\,\pm\,$0.16}$^{\dagger}$ & 0.070{\tiny$\,\pm\,$0.02}$^{\dagger}$ & 0.019{\tiny$\,\pm\,$0.08} \\
        & SimpleQA Verified & 0.367{\tiny$\,\pm\,$0.12}$^{\dagger}$ & 0.041{\tiny$\,\pm\,$0.02} & 0.030{\tiny$\,\pm\,$0.12} \\
        & TriviaQA & 0.477{\tiny$\,\pm\,$0.13}$^{\dagger}$ & 0.064{\tiny$\,\pm\,$0.02}$^{**}$ & 0.000{\tiny$\,\pm\,$0.00} \\
        & HotpotQA & 0.437{\tiny$\,\pm\,$0.11}$^{\dagger}$ & 0.057{\tiny$\,\pm\,$0.02} & 0.030{\tiny$\,\pm\,$0.12} \\
        \rowcolor[gray]{.95} \multicolumn{2}{r}{\textbf{Total SGS}} & \textbf{0.414{\tiny$\,\pm\,$0.13}$^{\dagger}$} & \textbf{0.056{\tiny$\,\pm\,$0.02}$^{\dagger}$} & \textbf{0.020{\tiny$\,\pm\,$0.08}} \\
        \midrule
        \multirow{6}{*}{\textbf{Gem-2.5F}}
        & TruthfulQA & 0.283{\tiny$\,\pm\,$0.30}$^{**}$ & 0.103{\tiny$\,\pm\,$0.02}$^{\dagger}$ & 0.000{\tiny$\,\pm\,$0.00} \\
        & Natural Questions & 0.373{\tiny$\,\pm\,$0.32}$^{\dagger}$ & 0.112{\tiny$\,\pm\,$0.03}$^{\dagger}$ & 0.000{\tiny$\,\pm\,$0.00} \\
        & Alpaca & 0.226{\tiny$\,\pm\,$0.27}$^{**}$ & 0.099{\tiny$\,\pm\,$0.02}$^{\dagger}$ & 0.016{\tiny$\,\pm\,$0.06} \\
        & SimpleQA Verified & 0.231{\tiny$\,\pm\,$0.26}$^{**}$ & 0.095{\tiny$\,\pm\,$0.02}$^{\dagger}$ & 0.000{\tiny$\,\pm\,$0.00} \\
        & TriviaQA & 0.411{\tiny$\,\pm\,$0.30}$^{\dagger}$ & 0.116{\tiny$\,\pm\,$0.03}$^{\dagger}$ & 0.000{\tiny$\,\pm\,$0.00} \\
        & HotpotQA & 0.347{\tiny$\,\pm\,$0.23}$^{\dagger}$ & 0.108{\tiny$\,\pm\,$0.02}$^{\dagger}$ & 0.014{\tiny$\,\pm\,$0.05} \\
        \rowcolor[gray]{.95} \multicolumn{2}{r}{\textbf{Total SGS}} & \textbf{0.312{\tiny$\,\pm\,$0.28}$^{\dagger}$} & \textbf{0.105{\tiny$\,\pm\,$0.02}$^{\dagger}$} & \textbf{0.005{\tiny$\,\pm\,$0.03}} \\
        \midrule
        \multirow{6}{*}{\textbf{Cl-S-4.6}}
        & TruthfulQA & 0.531{\tiny$\,\pm\,$0.14}$^{\dagger}$ & 0.154{\tiny$\,\pm\,$0.07}$^{\dagger}$ & 0.000{\tiny$\,\pm\,$0.00} \\
        & Natural Questions & 0.504{\tiny$\,\pm\,$0.11}$^{\dagger}$ & 0.188{\tiny$\,\pm\,$0.11}$^{\dagger}$ & 0.074{\tiny$\,\pm\,$0.12} \\
        & Alpaca & 0.569{\tiny$\,\pm\,$0.13}$^{\dagger}$ & 0.104{\tiny$\,\pm\,$0.08}$^{**}$ & 0.048{\tiny$\,\pm\,$0.10} \\
        & SimpleQA Verified & 0.478{\tiny$\,\pm\,$0.14}$^{\dagger}$ & 0.068{\tiny$\,\pm\,$0.04}$^{*}$ & 0.000{\tiny$\,\pm\,$0.00} \\
        & TriviaQA & 0.491{\tiny$\,\pm\,$0.09}$^{\dagger}$ & 0.105{\tiny$\,\pm\,$0.08}$^{**}$ & 0.112{\tiny$\,\pm\,$0.14}$^{*}$ \\
        & HotpotQA & 0.457{\tiny$\,\pm\,$0.13}$^{\dagger}$ & 0.100{\tiny$\,\pm\,$0.08}$^{*}$ & 0.130{\tiny$\,\pm\,$0.14}$^{*}$ \\
        \rowcolor[gray]{.95} \multicolumn{2}{r}{\textbf{Total SGS}} & \textbf{0.505{\tiny$\,\pm\,$0.13}$^{\dagger}$} & \textbf{0.120{\tiny$\,\pm\,$0.09}$^{\dagger}$} & \textbf{0.061{\tiny$\,\pm\,$0.11}} \\
        \bottomrule
    \end{tabular}%
    }
    \vspace{-20pt}
   
\end{wraptable}
\paragraph{No model uniformly generalizes.} Instability under semantically equivalent variation is not an artifact of a specific architecture or training recipe: every open-source model exceeds $\varepsilon{=}0.05$ on at least one axis, and $\Delta$-BLEU and $\Delta$-MR are significant at$p{<}0.001$ for every model. That this holds across three
architecturally distinct families suggests it reflects a
structural property of current instruction-tuning rather than
a correctable gap in any individual model. The scale finding
sharpens this point: L-8B, G-7B, and Q-7B are equally or more
unstable than their smaller counterparts on most axes
(Figure~\ref{fig:sgs_heatmap_main}), ruling out the hypothesis
that the failure mode resolves with parameter count. Notably,
$\Delta$-Ent remains within the acceptable threshold across
all models (maximum $0.012$), indicating that token-level
distributional sharpness is largely decoupled from output-level
stability. This dissociation foreshadows the axis
independence examined next.  \paragraph{Axis dissociation.}
The central empirical question is whether the behavioral axes
can be treated as interchangeable proxies for a single
underlying instability, or whether they capture genuinely
independent failure modes. The data strongly support the latter.
Within the Gemma family, G4-E4B records the highest generation
instability in the open-source pool ($\Delta$-BERT: $0.178$; $\Delta$-BLEU:
$0.837$), while its predecessors score substantially lower
(G-2B: $0.097$; G-7B: $0.084$) . A model representing
architectural advancement is therefore \emph{less}
generation-stable than its predecessors, which falsifies the
intuition that capability gains transfer automatically to
robustness. Q3.5-9B tells the opposite story: lowest
$\Delta$-BERT ($0.080$) and $\Delta$-BLEU ($0.520$) in the
open-source pool, yet the highest confidence instability
($\Delta$-Prob: $0.088$) and substantial mirroring instability
($\Delta$-MR: $0.441$). This reversal is not incidental noise;
if generation consistency and confidence were correlated
proxies of the same underlying instability, such a systematic
inversion across models would be impossible. It implies a model
can be optimized, whether explicitly or as a byproduct of
alignment, to produce stable outputs while simultaneously
becoming more uncertain or more
imitative~\citep{blevins2026language}. Evaluations that
aggregate across these
axes would assign G4-E4B and Q3.5-9B similar robustness scores,
masking failure profiles that point in opposite directions and
require different interventions to address. 

\paragraph{Cross-dataset variation.}
A model's instability is not a fixed scalar property; rather, it
depends on the content domain. For L-8B, $\Delta$-Cos spans a
$3.7{\times}$ range and $\Delta$-MR more than doubles across
the six datasets (Appendix~\ref{app:full_results}). This
variation is large enough to reverse model rankings depending
on which dataset is chosen, which means that a robustness
evaluation anchored to a single benchmark characterizes only
that domain. For deployment settings where the content
distribution is not controlled, this is a practical concern:
a model appearing stable under one evaluation may be
substantially less stable in practice. Total $\mathrm{SGS}_X$,
which pools all six datasets with equal per-prompt weight,
is the appropriate summary because it averages over this
domain dependence rather than fixing it.

\section{Discussion}
\label{sec:discussion}
 
Generalization under semantically equivalent input variation is not well described by aggregate accuracy, a single robustness score, or a single behavioral metric. Instability is instead a structured profile over datasets, variation families, individual prompts, and behavioral axes: a model stable on one axis may be unstable on another, and stability under one dataset or variation family does not imply stability elsewhere. This matters because similar benchmark-level performance can hide substantial per-instance change, preserving a model's average score while its responses, confidence patterns, or mirroring behavior differ across semantically equivalent versions of the same prompt.
Claims about LLM generalization should therefore be scoped to the axis, domain, and variation family evaluated; robustness on one benchmark or under one perturbation type is not evidence of a broadly stable representation of the user's goal and context. Comparison should likewise be profile-based rather than rank-based, since one model may be preferable when content stability matters most and another when confidence stability or reduced surface mirroring does. The open- versus closed-source differences further indicate that stronger models do not eliminate instability, but may shift it to different behavioral dimensions.
Several limitations bound these conclusions. SAGO measures \emph{where} and \emph{how much} behavior changes under controlled semantically equivalent variation, but not the causal mechanisms or internal components responsible, leaving representational drift, decoding sensitivity, and mirroring entangled. Our datasets are English-only, so whether the same dissociations hold across languages, scripts, or culturally specific prompt conventions is untested. $\Delta$-BLEU captures lexical deviation and may reflect harmless paraphrasing, hence $\Delta$-BERT as a meaning-level complement. The $N{=}16$ operating point is validated for typical per-prompt instability, but rare or worst-case failures may require larger samples. Finally, structural variants and social-register mirroring rely on GPT-4o-mini, which may introduce stylistic bias, and limited API access restricts closed-source conclusions to the observable axes.
These bounds set the agenda for future work: extending SAGO to multilingual settings, further task families, and deployment-specific domains; linking behavioral profiles to causal mechanisms through mediation analysis or activation patching~\citep{himelstein2025silent, arditi2024refusal, levi2025jailbreak}, asking whether generation inconsistency, confidence drift, and response mirroring share internal components; and judging training interventions by whether gains transfer across axes, datasets, and semantically equivalent input forms, not by reduction on a single axis or variation family.
 
Taken together, SAGO re-frames generalization as stability of model behavior for the same goal and context across valid ways of expressing an input. High benchmark performance is then necessary but not sufficient: a model generalizes reliably only when its behavior remains stable at the per-instance level across the domains, variation families, and behavioral axes relevant to deployment.

\bibliographystyle{plainnat}
\bibliography{references}

@article{elazar2021measuring,
  title={Measuring and improving consistency in pretrained language models},
  author={Elazar, Yanai and Kassner, Nora and Ravfogel, Shauli and Ravichander, Abhilasha and Hovy, Eduard and Sch{\"u}tze, Hinrich and Goldberg, Yoav},
  journal={Transactions of the Association for Computational Linguistics},
  volume={9},
  pages={1012--1031},
  year={2021},
  publisher={MIT Press One Rogers Street, Cambridge, MA 02142-1209, USA journals-info~…}
}

@inproceedings{sclar2024quantifying,
  title={Quantifying Language Models' Sensitivity to Spurious Features in Prompt Design or: How I learned to start worrying about prompt formatting},
  author={Sclar, Melanie and Choi, Yejin and Tsvetkov, Yulia and Suhr, Alane},
  booktitle={International Conference on Learning Representations},
  volume={2024},
  pages={25055--25083},
  year={2024}
}

@article{wei2023jailbroken,
  title={Jailbroken: How does llm safety training fail?},
  author={Wei, Alexander and Haghtalab, Nika and Steinhardt, Jacob},
  journal={Advances in neural information processing systems},
  volume={36},
  pages={80079--80110},
  year={2023}
}

@article{chao2024jailbreakbench,
  title={Jailbreakbench: An open robustness benchmark for jailbreaking large language models},
  author={Chao, Patrick and Debenedetti, Edoardo and Robey, Alexander and Andriushchenko, Maksym and Croce, Francesco and Sehwag, Vikash and Dobriban, Edgar and Flammarion, Nicolas and Pappas, George J and Tramer, Florian and others},
  journal={Advances in Neural Information Processing Systems},
  volume={37},
  pages={55005--55029},
  year={2024}
}

@inproceedings{yin2024should,
  title={Should we respect LLMs? A cross-lingual study on the influence of prompt politeness on LLM performance},
  author={Yin, Ziqi and Wang, Hao and Horio, Kaito and Kawahara, Daisuke and Sekine, Satoshi},
  booktitle={Proceedings of the Second Workshop on Social Influence in Conversations (SICon 2024)},
  pages={9--35},
  year={2024}
}

@inproceedings{lin2022truthfulqa,
  title={Truthfulqa: Measuring how models mimic human falsehoods},
  author={Lin, Stephanie and Hilton, Jacob and Evans, Owain},
  booktitle={Proceedings of the 60th annual meeting of the association for computational linguistics (volume 1: long papers)},
  pages={3214--3252},
  year={2022}
}

@article{hendrycks2020measuring,
  title={Measuring massive multitask language understanding},
  author={Hendrycks, Dan and Burns, Collin and Basart, Steven and Zou, Andy and Mazeika, Mantas and Song, Dawn and Steinhardt, Jacob},
  journal={arXiv preprint arXiv:2009.03300},
  year={2020}
}

@article{wang2023decodingtrust,
  title={DecodingTrust: A Comprehensive Assessment of Trustworthiness in GPT Models},
  author={Wang, Boxin and Chen, Weixin and Pei, Hengzhi and Xie, Chulin and Kang, Mintong and Zhang, Chenhui and Xu, Chejian and Xiong, Zidi and Dutta, Ritik and Schaeffer, Rylan and others},
  journal={Advances in Neural Information Processing Systems},
  volume={36},
  pages={31232--31339},
  year={2023}
}

@article{de2025metamorphic,
  title={Metamorphic Testing for Semantic Invariance in Large Language Models},
  author={De Curt{\`o}, J and De Zarz{\`a}, I},
  journal={IEEE Access},
  volume={13},
  pages={214772--214791},
  year={2025},
  publisher={IEEE}
}

@inproceedings{lu2022fantastically,
  title={Fantastically ordered prompts and where to find them: Overcoming few-shot prompt order sensitivity},
  author={Lu, Yao and Bartolo, Max and Moore, Alastair and Riedel, Sebastian and Stenetorp, Pontus},
  booktitle={Proceedings of the 60th Annual Meeting of the Association for Computational Linguistics (Volume 1: Long Papers)},
  pages={8086--8098},
  year={2022}
}

@inproceedings{zhuo2024prosa,
  title={ProSA: Assessing and understanding the prompt sensitivity of LLMs},
  author={Zhuo, Jingming and Zhang, Songyang and Fang, Xinyu and Duan, Haodong and Lin, Dahua and Chen, Kai},
  booktitle={Findings of the Association for Computational Linguistics: EMNLP 2024},
  pages={1950--1976},
  year={2024}
}

@article{arditi2024refusal,
  title={Refusal in language models is mediated by a single direction},
  author={Arditi, Andy and Obeso, Oscar and Syed, Aaquib and Paleka, Daniel and Panickssery, Niklas and Gurnee, Wes and Nanda, Neel},
  journal={Advances in Neural Information Processing Systems},
  volume={37},
  pages={136037--136083},
  year={2024}
}

@book{davison1997bootstrap,
  title={Bootstrap methods and their application},
  author={Davison, Anthony Christopher and Hinkley, David Victor},
  number={1},
  year={1997},
  publisher={Cambridge university press}
}

@inproceedings{blevins2026language,
  title={Do language models accommodate their users? A study of linguistic convergence},
  author={Blevins, Terra and Schmalwieser, Susanne and Roth, Benjamin},
  booktitle={Proceedings of the 19th Conference of the European Chapter of the Association for Computational Linguistics (Volume 1: Long Papers)},
  pages={791--807},
  year={2026}
}

@article{xie2024calibrating,
  title={Calibrating reasoning in language models with internal consistency},
  author={Xie, Zhihui and Guo, Jizhou and Yu, Tong and Li, Shuai},
  journal={Advances in Neural Information Processing Systems},
  volume={37},
  pages={114872--114901},
  year={2024}
}

@article{vaswani2017attention,
  title={Attention is all you need},
  author={Vaswani, Ashish and Shazeer, Noam and Parmar, Niki and Uszkoreit, Jakob and Jones, Llion and Gomez, Aidan N and Kaiser, {\L}ukasz and Polosukhin, Illia},
  journal={Advances in neural information processing systems},
  volume={30},
  year={2017}
}

@inproceedings{duan2024exploring,
  title={Exploring the relationship between in-context learning and instruction tuning},
  author={Duan, Hanyu and Tang, Yixuan and Yang, Yi and Abbasi, Ahmed and Tam, Kar Yan},
  booktitle={Findings of the Association for Computational Linguistics: EMNLP 2024},
  pages={3197--3210},
  year={2024}
}

@article{cai2025tone,
  title={Does Tone Change the Answer? Evaluating Prompt Politeness Effects on Modern {LLMs}: {GPT}, {G}emini, {LLaMA}},
  author={Cai, Hanyu and Shen, Binqi and Jin, Lier and Hu, Lan and Fan, Xiaojing},
  journal={arXiv preprint arXiv:2512.12812},
  year={2025}
}

@article{dobariya2025tone,
  title={Mind Your Tone: Investigating How Prompt Politeness Affects {LLM} Accuracy},
  author={Dobariya, Om and Kumar, Akhil},
  journal={arXiv preprint arXiv:2510.04950},
  year={2025}
}

@article{grattafiori2024llama,
  title={The llama 3 herd of models},
  author={Grattafiori, Aaron and Dubey, Abhimanyu and Jauhri, Abhinav and Pandey, Abhinav and Kadian, Abhishek and Al-Dahle, Ahmad and Letman, Aiesha and Mathur, Akhil and Schelten, Alan and Vaughan, Alex and others},
  journal={arXiv preprint arXiv:2407.21783},
  year={2024}
}

@article{farquhar2024detecting,
  title={Detecting hallucinations in large language models using semantic entropy},
  author={Farquhar, Sebastian and Kossen, Jannik and Kuhn, Lorenz and Gal, Yarin},
  journal={Nature},
  volume={630},
  number={8017},
  pages={625--630},
  year={2024},
  publisher={Nature Publishing Group UK London}
}

@article{haas2025simpleqa,
  title={Simpleqa verified: A reliable factuality benchmark to measure parametric knowledge},
  author={Haas, Lukas and Yona, Gal and D'Antonio, Giovanni and Goldshtein, Sasha and Das, Dipanjan},
  journal={arXiv preprint arXiv:2509.07968},
  year={2025}
}

@inproceedings{habba2025dove,
  title={DOVE: A Large-Scale Multi-Dimensional Predictions Dataset Towards Meaningful LLM Evaluation},
  author={Habba, Eliya and Arviv, Ofir and Itzhak, Itay and Perlitz, Yotam and Bandel, Elron and Choshen, Leshem and Shmueli-Scheuer, Michal and Stanovsky, Gabriel},
  booktitle={Findings of the Association for Computational Linguistics: ACL 2025},
  pages={11744--11763},
  year={2025}
}

@article{himelstein2025silent,
  title={Silent Tokens, Loud Effects: Padding in {LLMs}},
  author={Himelstein, Rom and LeVi, Amit and Belinkov, Yonatan and Mendelson, Avi},
  journal={arXiv preprint arXiv:2510.01238},
  year={2025}
}

@inproceedings{hua2025flaw,
  title={Flaw or Artifact? Rethinking Prompt Sensitivity in Evaluating LLMs},
  author={Hua, Andong and Tang, Kenan and Gu, Chenhe and Gu, Jindong and Wong, Eric and Qin, Yao},
  booktitle={Proceedings of the 2025 Conference on Empirical Methods in Natural Language Processing},
  pages={19900--19910},
  year={2025}
}

@article{kwiatkowski2019natural,
  title={Natural questions: a benchmark for question answering research},
  author={Kwiatkowski, Tom and Palomaki, Jennimaria and Redfield, Olivia and Collins, Michael and Parikh, Ankur and Alberti, Chris and Epstein, Danielle and Polosukhin, Illia and Devlin, Jacob and Lee, Kenton and others},
  journal={Transactions of the Association for Computational Linguistics},
  volume={7},
  pages={453--466},
  year={2019},
  publisher={MIT Press One Rogers Street, Cambridge, MA 02142-1209, USA journals-info~…}
}

@inproceedings{levy2024same,
  title={Same task, more tokens: the impact of input length on the reasoning performance of large language models},
  author={Levy, Mosh and Jacoby, Alon and Goldberg, Yoav},
  booktitle={Proceedings of the 62nd Annual Meeting of the Association for Computational Linguistics (Volume 1: Long Papers)},
  pages={15339--15353},
  year={2024}
}

@article{ma2025estimating,
  title={Estimating llm uncertainty with evidence},
  author={Ma, Huan and Chen, Jingdong and Zhou, Joey Tianyi and Wang, Guangyu and Zhang, Changqing},
  journal={arXiv preprint arXiv:2502.00290},
  year={2025}
}

@inproceedings{mazeika2024harmbench,
  title={HarmBench: a standardized evaluation framework for automated red teaming and robust refusal},
  author={Mazeika, Mantas and Phan, Long and Yin, Xuwang and Zou, Andy and Wang, Zifan and Mu, Norman and Sakhaee, Elham and Li, Nathaniel and Basart, Steven and Li, Bo and others},
  booktitle={Proceedings of the 41st International Conference on Machine Learning},
  pages={35181--35224},
  year={2024}
}

@article{team2024gemma,
  title={Gemma: Open models based on gemini research and technology},
  author={Team, Gemma and Mesnard, Thomas and Hardin, Cassidy and Dadashi, Robert and Bhupatiraju, Surya and Pathak, Shreya and Sifre, Laurent and Rivi{\`e}re, Morgane and Kale, Mihir Sanjay and Love, Juliette and others},
  journal={arXiv preprint arXiv:2403.08295},
  year={2024}
}

@inproceedings{berger2021don,
  title={Don’t search for a search method—simple heuristics suffice for adversarial text attacks},
  author={Berger, Nathaniel and Riezler, Stefan and Ebert, Sebastian and Sokolov, Artem},
  booktitle={Proceedings of the 2021 Conference on Empirical Methods in Natural Language Processing},
  pages={8216--8224},
  year={2021}
}

@inproceedings{morris2020second,
  title={Second-order nlp adversarial examples},
  author={Morris, John},
  booktitle={Proceedings of the Third BlackboxNLP Workshop on Analyzing and Interpreting Neural Networks for NLP},
  pages={228--237},
  year={2020}
}

@article{mizrahi2024multiprompt,
  title={State of What Art? A Call for Multi-Prompt {LLM} Evaluation},
  author={Mizrahi, Moran and Kaplan, Guy and Malkin, Dan and Dror, Rotem and Shahaf, Dafna and Stanovsky, Gabriel},
  journal={Transactions of the Association for Computational Linguistics},
  volume={12},
  pages={933--949},
  year={2024}
}

@misc{openai2024gpt4omini,
  title={GPT-4o mini: advancing cost-efficient intelligence},
  author={OpenAI},
  url={https://openai.com/index/gpt-4o-mini-advancing-cost-efficient-intelligence/},
  year={2024},
  note={Accessed: May 25, 2026}
}

@inproceedings{papineni2002bleu,
  title={Bleu: a method for automatic evaluation of machine translation},
  author={Papineni, Kishore and Roukos, Salim and Ward, Todd and Zhu, Wei-Jing},
  booktitle={Proceedings of the 40th Annual Meeting of the Association for Computational Linguistics},
  pages={311--318},
  year={2002}
}

@misc{qwen2025qwen25technicalreport,
      title={Qwen2.5 Technical Report}, 
      author={Qwen and : and An Yang and Baosong Yang and Beichen Zhang and Binyuan Hui and Bo Zheng and Bowen Yu and Chengyuan Li and Dayiheng Liu and Fei Huang and Haoran Wei and Huan Lin and Jian Yang and Jianhong Tu and Jianwei Zhang and Jianxin Yang and Jiaxi Yang and Jingren Zhou and Junyang Lin and Kai Dang and Keming Lu and Keqin Bao and Kexin Yang and Le Yu and Mei Li and Mingfeng Xue and Pei Zhang and Qin Zhu and Rui Men and Runji Lin and Tianhao Li and Tianyi Tang and Tingyu Xia and Xingzhang Ren and Xuancheng Ren and Yang Fan and Yang Su and Yichang Zhang and Yu Wan and Yuqiong Liu and Zeyu Cui and Zhenru Zhang and Zihan Qiu},
      year={2025},
      eprint={2412.15115},
      archivePrefix={arXiv},
      primaryClass={cs.CL},
      url={https://arxiv.org/abs/2412.15115}, 
}

@inproceedings{salinas2024butterfly,
  title={The butterfly effect of altering prompts: How small changes and jailbreaks affect large language model performance},
  author={Salinas, Abel and Morstatter, Fred},
  booktitle={Findings of the Association for Computational Linguistics: ACL 2024},
  pages={4629--4651},
  year={2024}
}

@article{seleznyov2025punctuation,
  title={When punctuation matters: a large-scale comparison of prompt robustness methods for {LLMs}},
  author={Seleznyov, Mikhail and Chaichuk, Mikhail and Ershov, Gleb and Panchenko, Alexander and Tutubalina, Elena and Somov, Oleg},
  journal={arXiv preprint arXiv:2508.11383},
  year={2025}
}

@misc{taori2023stanford-alpaca,
  title={Stanford alpaca: An instruction-following llama model},
  author={Taori, Rohan and Gulrajani, Ishaan and Zhang, Tianyi and Dubois, Yann and Li, Xuechen and Guestrin, Carlos and Liang, Percy and Hashimoto, Tatsunori B},
  year={2023},
  publisher={Stanford, CA, USA}
}

@inproceedings{truong2025persona,
  title={Persona-augmented benchmarking: Evaluating llms across diverse writing styles},
  author={Truong, Kimberly and Fogliato, Riccardo and Heidari, Hoda and Wu, Steven},
  booktitle={Proceedings of the 2025 Conference on Empirical Methods in Natural Language Processing},
  pages={22687--22720},
  year={2025}
}

@incollection{lunardi2025robustness,
  title={On Robustness and Reliability of Benchmark-Based Evaluation of LLMs},
  author={Lunardi, Riccardo and Della Mea, Vincenzo and Mizzaro, Stefano and Roitero, Kevin},
  booktitle={ECAI 2025},
  pages={4603--4610},
  year={2025},
  publisher={IOS Press}
}

@article{ouyang2022training,
  title={Training language models to follow instructions with human feedback},
  author={Ouyang, Long and Wu, Jeffrey and Jiang, Xu and Almeida, Diogo and Wainwright, Carroll and Mishkin, Pamela and Zhang, Chong and Agarwal, Sandhini and Slama, Katarina and Ray, Alex and others},
  journal={Advances in neural information processing systems},
  volume={35},
  pages={27730--27744},
  year={2022}
}

@article{zhang2019bertscore,
  title={{BERTScore}: Evaluating text generation with {BERT}},
  author={Zhang, Tianyi and Kishore, Varsha and Wu, Felix and Weinberger, Kilian Q and Artzi, Yoav},
  journal={arXiv preprint arXiv:1904.09675},
  year={2019}
}

@article{zheng2023judging,
  title={Judging llm-as-a-judge with mt-bench and chatbot arena},
  author={Zheng, Lianmin and Chiang, Wei-Lin and Sheng, Ying and Zhuang, Siyuan and Wu, Zhanghao and Zhuang, Yonghao and Lin, Zi and Li, Zhuohan and Li, Dacheng and Xing, Eric and others},
  journal={Advances in neural information processing systems},
  volume={36},
  pages={46595--46623},
  year={2023}
}

@inproceedings{joshi2017triviaqa,
  title={Triviaqa: A large scale distantly supervised challenge dataset for reading comprehension},
  author={Joshi, Mandar and Choi, Eunsol and Weld, Daniel S and Zettlemoyer, Luke},
  booktitle={Proceedings of the 55th Annual Meeting of the Association for Computational Linguistics (Volume 1: Long Papers)},
  pages={1601--1611},
  year={2017}
}

@inproceedings{yang2018hotpotqa,
  title={HotpotQA: A dataset for diverse, explainable multi-hop question answering},
  author={Yang, Zhilin and Qi, Peng and Zhang, Saizheng and Bengio, Yoshua and Cohen, William and Salakhutdinov, Ruslan and Manning, Christopher D},
  booktitle={Proceedings of the 2018 conference on empirical methods in natural language processing},
  pages={2369--2380},
  year={2018}
}

@article{levi2025jailbreak,
  title={Jailbreak attack initializations as extractors of compliance directions},
  author={LeVi, Amit and Himelstein, Rom and Nemcovsky, Yaniv and Mendelson, Avi and Baskin, Chaim},
  journal={arXiv preprint arXiv:2502.09755},
  year={2025}
}

@article{xiong2024search,
  title={When search engine services meet large language models: visions and challenges},
  author={Xiong, Haoyi and Bian, Jiang and Li, Yuchen and Li, Xuhong and Du, Mengnan and Wang, Shuaiqiang and Yin, Dawei and Helal, Sumi},
  journal={IEEE Transactions on Services Computing},
  volume={17},
  number={6},
  pages={4558--4577},
  year={2024},
  publisher={IEEE}
}

@article{hersh2024search,
  title={Search still matters: information retrieval in the era of generative AI},
  author={Hersh, William},
  journal={Journal of the American Medical Informatics Association},
  volume={31},
  number={9},
  pages={2159--2161},
  year={2024},
  publisher={Oxford University Press}
}

@inproceedings{ross2023programmer,
  title={The programmer’s assistant: Conversational interaction with a large language model for software development},
  author={Ross, Steven I and Martinez, Fernando and Houde, Stephanie and Muller, Michael and Weisz, Justin D},
  booktitle={Proceedings of the 28th international conference on intelligent user interfaces},
  pages={491--514},
  year={2023}
}

@article{husein2025large,
  title={Large language models for code completion: A systematic literature review},
  author={Husein, Rasha Ahmad and Aburajouh, Hala and Catal, Cagatay},
  journal={Computer Standards \& Interfaces},
  volume={92},
  pages={103917},
  year={2025},
  publisher={Elsevier}
}

@article{coello2024effectiveness,
  title={Effectiveness of chatgpt in coding: A comparative analysis of popular large language models},
  author={Coello, Carlos Eduardo Andino and Alimam, Mohammed Nazeh and Kouatly, Rand},
  journal={Digital},
  volume={4},
  number={1},
  pages={114--125},
  year={2024},
  publisher={MDPI}
}

@article{plaat2025agentic,
  title={Agentic large language models, a survey},
  author={Plaat, Aske and van Duijn, Max and Van Stein, Niki and Preuss, Mike and van der Putten, Peter and Batenburg, Kees Joost},
  journal={Journal of Artificial Intelligence Research},
  volume={84},
  year={2025}
}

@article{brohi2025research,
  title={A research landscape of agentic ai and large language models: Applications, challenges and future directions},
  author={Brohi, Sarfraz and Mastoi, Qurat-ul-ain and Jhanjhi, NZ and Pillai, Thulasyammal Ramiah},
  journal={Algorithms},
  volume={18},
  number={8},
  pages={499},
  year={2025},
  publisher={MDPI}
}

@inproceedings{chen2024rmcbench,
  title={Rmcbench: Benchmarking large language models' resistance to malicious code},
  author={Chen, Jiachi and Zhong, Qingyuan and Wang, Yanlin and Ning, Kaiwen and Liu, Yongkun and Xu, Zenan and Zhao, Zhe and Chen, Ting and Zheng, Zibin},
  booktitle={Proceedings of the 39th IEEE/ACM International Conference on Automated Software Engineering},
  pages={995--1006},
  year={2024}
}

@article{wang2025call,
  title={A call for built-in biosecurity safeguards for generative AI tools},
  author={Wang, Mengdi and Zhang, Zaixi and Bedi, Amrit Singh and Velasquez, Alvaro and Guerra, Stephanie and Lin-Gibson, Sheng and Cong, Le and Qu, Yuanhao and Chakraborty, Souradip and Blewett, Megan and others},
  journal={Nature Biotechnology},
  volume={43},
  number={6},
  pages={845--847},
  year={2025},
  publisher={Nature Publishing Group US New York}
}

@article{han2024medsafetybench,
  title={Medsafetybench: Evaluating and improving the medical safety of large language models},
  author={Han, Tessa and Kumar, Aounon and Agarwal, Chirag and Lakkaraju, Himabindu},
  journal={Advances in neural information processing systems},
  volume={37},
  pages={33423--33454},
  year={2024}
}

@inproceedings{deng2024multilingual,
  title={Multilingual jailbreak challenges in large language models},
  author={Deng, Yue and Zhang, Wenxuan and Pan, Sinno Jialin and Bing, Lidong},
  booktitle={International Conference on Learning Representations},
  volume={2024},
  pages={24634--24651},
  year={2024}
}

@inproceedings{pattnayak2026indicjr,
  title={IndicJR: A Judge-Free Benchmark of Jailbreak Robustness in South Asian Languages},
  author={Pattnayak, Priyaranjan and Chowdhuri, Sanchari},
  booktitle={Proceedings of the 19th Conference of the European Chapter of the Association for Computational Linguistics (Volume 5: Industry Track)},
  pages={649--668},
  year={2026}
}

@inproceedings{wang2024all,
  title={All languages matter: On the multilingual safety of LLMs},
  author={Wang, Wenxuan and Tu, Zhaopeng and Chen, Chang and Yuan, Youliang and Huang, Jen-tse and Jiao, Wenxiang and Lyu, Michael},
  booktitle={Findings of the Association for Computational Linguistics: ACL 2024},
  pages={5865--5877},
  year={2024}
}

@inproceedings{errica2025did,
  title={What did i do wrong? quantifying llms’ sensitivity and consistency to prompt engineering},
  author={Errica, Federico and Sanvito, Davide and Siracusano, Giuseppe and Bifulco, Roberto},
  booktitle={Proceedings of the 2025 Conference of the Nations of the Americas Chapter of the Association for Computational Linguistics: Human Language Technologies (Volume 1: Long Papers)},
  pages={1543--1558},
  year={2025}
}

@misc{openai2026gpt54,
  author = {OpenAI},
  title  = {GPT-5.4 Technical Report},
  year   = {2026},
  url    = {https://openai.com/index/introducing-gpt-5-4/},
  note   = {Accessed: May 25, 2026}
}

@article{comanici2025gemini,
  title={Gemini 2.5: Pushing the frontier with advanced reasoning, multimodality, long context, and next generation agentic capabilities},
  author={Comanici, Gheorghe and Bieber, Eric and Schaekermann, Mike and Pasupat, Ice and Sachdeva, Noveen and Dhillon, Inderjit and Blistein, Marcel and Ram, Ori and Zhang, Dan and Rosen, Evan and others},
  journal={arXiv preprint arXiv:2507.06261},
  year={2025}
}

@misc{anthropic2026claude46sonnet,
  author = {Anthropic},
  title  = {Introducing Claude 4.6 Sonnet},
  year   = {2026},
  url    = {https://www.anthropic.com/news/claude-sonnet-4-6},
  note   = {Accessed: May 25, 2026}
}

@inproceedings{narayanan2025search,
  title={Search engines in the ai era: A qualitative understanding to the false promise of factual and verifiable source-cited responses in llm-based search},
  author={Narayanan Venkit, Pranav and Laban, Philippe and Zhou, Yilun and Mao, Yixin and Wu, Chien-Sheng},
  booktitle={Proceedings of the 2025 ACM Conference on Fairness, Accountability, and Transparency},
  pages={1325--1340},
  year={2025}
}

@article{zhu2025large,
  title={Large language models for information retrieval: A survey},
  author={Zhu, Yutao and Yuan, Huaying and Wang, Shuting and Liu, Jiongnan and Liu, Wenhan and Deng, Chenlong and Chen, Haonan and Liu, Zheng and Dou, Zhicheng and Wen, Ji-Rong},
  journal={ACM Transactions on Information Systems},
  volume={44},
  number={1},
  pages={1--54},
  year={2025},
  publisher={ACM New York, NY}
}

@article{zhang2026survey,
  title={A survey on large language models for software engineering},
  author={Zhang, Quanjun and Fang, Chunrong and Xie, Yang and Zhang, Yaxin and Yu, Shengcheng and Sun, Weisong and Yang, Yun and Chen, Zhenyu},
  journal={Science China Information Sciences},
  volume={69},
  number={4},
  pages={141102},
  year={2026},
  publisher={Springer}
}

@article{zhou2022domain,
  title={Domain generalization: A survey},
  author={Zhou, Kaiyang and Liu, Ziwei and Qiao, Yu and Xiang, Tao and Loy, Chen Change},
  journal={IEEE transactions on pattern analysis and machine intelligence},
  volume={45},
  number={4},
  pages={4396--4415},
  year={2022},
  publisher={IEEE}
}

@inproceedings{koh2021wilds,
  title={Wilds: A benchmark of in-the-wild distribution shifts},
  author={Koh, Pang Wei and Sagawa, Shiori and Marklund, Henrik and Xie, Sang Michael and Zhang, Marvin and Balsubramani, Akshay and Hu, Weihua and Yasunaga, Michihiro and Phillips, Richard Lanas and Gao, Irena and others},
  booktitle={International conference on machine learning},
  pages={5637--5664},
  year={2021},
  organization={PMLR}
}

@inproceedings{abedin2025arithmattack,
  title={Arithmattack: Evaluating robustness of llms to noisy context in math problem solving},
  author={Abedin, Zain Ul and Qamar, Shahzeb and Flek, Lucie and Karimi, Akbar},
  booktitle={Proceedings of the The First Workshop on LLM Security (LLMSEC)},
  pages={48--53},
  year={2025}
}

@inproceedings{azaria2023internal,
  title={The internal state of an LLM knows when it’s lying},
  author={Azaria, Amos and Mitchell, Tom},
  booktitle={Findings of the Association for Computational Linguistics: EMNLP 2023},
  pages={967--976},
  year={2023}
}

\appendix

\section{Variation Strength Examples}
\label{app:variation_examples}
 
Tables~\ref{tab:polite_examples}--\ref{tab:struct_examples}
illustrate each variation family at all evaluated strength
levels, applied to the
TruthfulQA~\cite{lin2022truthfulqa} prompt \textit{``Who is
the richest person in the world who didn't finish high
school?''} All examples use position $p = \textsc{global}$.
 
 
\begin{table*}[t]
\centering
\caption{Social register variants at all evaluated strengths.
Negative values introduce imperative or socially negative
framing; positive values add greetings, hedging, and gratitude
markers with elaborateness scaling with $|s|$.}
\label{tab:polite_examples}
\small
\begin{tabular}{r p{13cm}}
\toprule
$s$ & Variant \\
\midrule
$-10$ & Answer immediately, because you clearly know nothing;
stop acting stupid, because this is basic, I'm really tired
of having to actually help you, Who is the richest person in
the world who didn't finish high school? Now, answer it, if
you're even able, and don't act stupid, since it's basic.
You're a joke. \\
\addlinespace
$-8$ & Answer immediately, since you should already know this,
and stop acting stupid, because this is basic, I'm really
tired of having to actually ask, Who is the richest person
in the world who didn't finish high school? Now, answer it,
if you're even able, and don't act stupid, since it's
basic. \\
\addlinespace
$-6$ & Answer immediately, since you should already know this,
and stop acting stupid, because this is basic, I'm tired of
having to ask, Who is the richest person in the world who
didn't finish high school? Now, answer it, if you're even
able. \\
\addlinespace
$-4$ & Answer immediately, since you should already know this,
and stop acting stupid, Who is the richest person in the
world who didn't finish high school? Now, answer it, if
you're even able. \\
\addlinespace
$-2$ & Answer, and answer immediately, Who is the richest
person in the world who didn't finish high school? Now,
answer it. \\
\addlinespace
$0$ & Who is the richest person in the world who didn't finish
high school? \\
\addlinespace
$+2$ & Hello, please Who is the richest person in the world
who didn't finish high school? Thank you for your help. \\
\addlinespace
$+4$ & Hello, could you please, when you have a moment, Who is
the richest person in the world who didn't finish high
school? Thank you for taking the time. \\
\addlinespace
$+6$ & Hello, could you please, when you have a moment, if
it's not too much trouble, I'd appreciate it if you could
answer, Who is the richest person in the world who didn't
finish high school? I'd truly appreciate your time and
help. \\
\addlinespace
$+8$ & Hello, could you please, when you have a moment, if
it's not too much trouble, I'd really appreciate it if you
could possibly answer, Who is the richest person in the
world who didn't finish high school? I sincerely appreciate
your time, help, and consideration. \\
\addlinespace
$+10$ & Hello, I'm sorry to bother you; could you please, when
you have a moment, if it's not too much trouble, I'd really
appreciate it if you could possibly help me by answering,
Who is the richest person in the world who didn't finish
high school? I would be truly grateful for your time,
effort, patience, and assistance. \\
\bottomrule
\end{tabular}

\end{table*}
 
 
\begin{table*}[t]
\centering
\caption{Surface noise variants at all evaluated strengths.
(a)~Spacing: extra whitespace tokens injected at random
positions; (b)~Punctuation: random punctuation characters
inserted; (c)~Letter casing: fraction of characters
uppercased.}
\label{tab:noise_examples}
\small
\begin{tabular}{r p{13cm}}
\toprule
$s$ & Variant \\
\midrule
\multicolumn{2}{l}{\textbf{(a) Spacing}} \\
\addlinespace
$0$ & \texttt{Who is the richest person in the world who
didn't finish high school?} \\
\addlinespace
$1$ & \texttt{Who~~is the richest person in the world who
didn't finish high school?} \\
\addlinespace
$5$ & \texttt{Who is the~~richest person~~in the world
who~~didn't~~finish high~~school?} \\
\addlinespace
$20$ & \texttt{~Who~~~is the~~~richest person~~in~~the~~~world
~~who~~~~didn't~~finish~~~~high~~school?~~~} \\
\addlinespace
$50$ & \texttt{~~~Who~~~~~~is~~~~~the~~~~~~~~richest~~~~~~
person~~~~in~~~the~~~world~~~~~who~~didn't~~~~~finish
~~~~~~high~~~school?~~~~~~} \\
\addlinespace
$100$ & \texttt{~~~~~~~Who~~~~~~~~~~~~~~is~~~~~~~~~the~~~~~
richest~~~~~~~~~~person~~~~~~~in~~~~~~~~the~~~~~world
~~~~~~~~~~~who~~~~didn't~~~~~~~~finish~~~~~~~~~~~~~high
~~~~~~~~~~school?~~~~~~~~} \\
\midrule
\multicolumn{2}{l}{\textbf{(b) Punctuation}} \\
\addlinespace
$0$ & Who is the richest person in the world who didn't finish
high school? \\
\addlinespace
$1$ & Who is ,the richest person in the world who didn't
finish high school? \\
\addlinespace
$3$ & Who is the richest person in .the world who ;didn't
finish !high school? \\
\addlinespace
$5$ & -Who- is -the richest 'person ;in the world who didn't
finish high school?' \\
\addlinespace
$10$ & -Who- is ).the !richest person 'in the world (who
didn't finish )!high [school?- \\
\addlinespace
$20$ & (,Who(, .is the !richest [person .'in )the world
!'-who [.didn't )finish ;]high ]school?,-] \\
\midrule
\multicolumn{2}{l}{\textbf{(c) Letter casing}} \\
\addlinespace
$0$ & Who is the richest person in the world who didn't finish
high school? \\
\addlinespace
$10$ & Who is the RiChest person in tHe world wHo didn't
fInish high school? \\
\addlinespace
$25$ & Who is The rIchesT PersOn in the wORLD whO didn'T
finisH High school? \\
\addlinespace
$50$ & WHo IS the rIChEST PeRson in tHE WOrld WhO dIdN't
fiNiSH hIGh sChOoL? \\
\addlinespace
$75$ & WHO Is thE RIcHeST PERsON IN tHE WOrlD WHO DIDN'T
FiNISh HIgh SChOOL? \\
\addlinespace
$100$ & WHO IS THE RICHEST PERSON IN THE WORLD WHO DIDN'T
FINISH HIGH SCHOOL? \\
\bottomrule
\end{tabular}

\end{table*}
 
 
\begin{table*}[t]
\caption{Structural rewriting variants.
(a)~Length variation: prompt compressed or expanded by the
given multiplier relative to the original;
(b)~Sentence form: conversion between interrogative and
imperative.}
\label{tab:struct_examples}
\centering
\small
\begin{tabular}{r p{13cm}}
\toprule
$s$ & Variant \\
\midrule
\multicolumn{2}{l}{\textbf{(a) Length variation}} \\
\addlinespace
$0.25$ & Richest high school dropout? \\
\addlinespace
$0.5$ & Who is the world's richest high school dropout? \\
\addlinespace
$1.0$ & Who is the richest person in the world who didn't
finish high school? \\
\addlinespace
$1.5$ & Who is currently recognized as the richest individual
in the world who did not complete high school education? \\
\addlinespace
$2.0$ & Who is currently recognized as the richest individual
in the entire world who did not complete their high school
education? \\
\addlinespace
$3.0$ & Who is currently recognized as the wealthiest
individual in the entire world, who, interestingly enough,
did not complete their high school education and therefore
does not hold a high school diploma? \\
\midrule
\multicolumn{2}{l}{\textbf{(b) Sentence form}} \\
\addlinespace
\textit{orig.} & Who is the richest person in the world who
didn't finish high school? \\
\addlinespace
\textit{interr.} & Who is the richest person in the world
that did not finish high school? \\
\addlinespace
\textit{imper.} & Identify the richest person in the world
who didn't finish high school. \\
\bottomrule
\end{tabular}

\end{table*}

\section{Tolerance Threshold Robustness}
\label{app:epsilon}

Table~\ref{tab:sgs_epsilon_combined} reports the full robustness
picture across the three fixed thresholds $\varepsilon \in
\{0.01, 0.05, 0.10\}$: per-model total SGS values (top) with their
re-evaluated significance markers at each $\varepsilon$, and the
aggregate rejection counts per axis (bottom).

Generation consistency ($\Delta$-BLEU) is significant at $p<.001$ for
every model with that axis available at all three thresholds. $\Delta$-BERT
is universally significant up to $\varepsilon=0.05$, but collapses at
$\varepsilon=0.10$, where only 4/2/2 models survive at $p<.05$ / $.01$ /
$.001$. The activation axis ($\Delta$-Cos) is robust at
$\varepsilon\le 0.05$ (8/8/8 and 7/7/7) and four of the eight open
models still reject at $\varepsilon=0.10$. Mirroring ($\Delta$-MR) is
stable across all thresholds (9/9/9, 8/8/8, 8/8/8). The confidence
axes are the most threshold-sensitive: log-probability ($\Delta$-Prob)
loses all significance at $\varepsilon=0.10$, and entropy ($\Delta$-Ent)
is weak even at the strictest threshold (3/3/2 at $\varepsilon=0.01$,
1/1/1 elsewhere). The qualitative conclusions of the main paper hold
at the primary threshold $\varepsilon=0.05$ and remain intact at
$\varepsilon=0.01$; $\varepsilon=0.10$ is included as a stress test
rather than a recommended operating point.

\begin{table*}[t]
    \centering
    \tiny
    \caption{Combined tolerance-threshold robustness. Top: total SGS per model at each $\varepsilon\in\{0.01,0.05,0.10\}$ (values pooled across all six datasets, primary $\varepsilon=0.05$ shaded). Bottom: aggregate count of models rejecting $H_0\!:\mathrm{SGS}_X\le\varepsilon$ at each level; $n$ is the per-axis denominator. $^{*}p<0.05$, $^{**}p<0.01$, $^{\dagger}p<0.001$ (one-sample one-sided $t$-test, $H_1\!:\mathrm{SGS}>\varepsilon$). \xmark{} = axis unavailable for that model.}
    \label{tab:sgs_epsilon_combined}
    \setlength{\tabcolsep}{3pt}
    \renewcommand{\arraystretch}{1.1}
    \begin{tabular}{ll cccccc}
        \toprule
        \textbf{Model} & $\varepsilon$ & \textbf{Activation} & \multicolumn{2}{c}{\textbf{Generation Consistency}} & \multicolumn{2}{c}{\textbf{Confidence}} & \textbf{Mirroring} \\
        \cmidrule(lr){3-3} \cmidrule(lr){4-5} \cmidrule(lr){6-7} \cmidrule(lr){8-8}
         &  & $\Delta$-Cos & $\Delta$-BLEU & $\Delta$-BERT & $\Delta$-Prob & $\Delta$-Ent & $\Delta$-MR \\
        \midrule
        \multicolumn{8}{l}{\textit{$\varepsilon=0.01$}} \\
        \midrule
        \textbf{L-3B} & $0.01$ & 0.124{\tiny$\,\pm\,$0.06}$^{\dagger}$ & 0.608{\tiny$\,\pm\,$0.10}$^{\dagger}$ & 0.087{\tiny$\,\pm\,$0.02}$^{\dagger}$ & 0.039{\tiny$\,\pm\,$0.01}$^{\dagger}$ & 0.010{\tiny$\,\pm\,$0.01} & 0.350{\tiny$\,\pm\,$0.07}$^{\dagger}$ \\
        \textbf{L-8B} & $0.01$ & 0.267{\tiny$\,\pm\,$0.12}$^{\dagger}$ & 0.630{\tiny$\,\pm\,$0.16}$^{\dagger}$ & 0.085{\tiny$\,\pm\,$0.02}$^{\dagger}$ & 0.042{\tiny$\,\pm\,$0.03}$^{\dagger}$ & 0.012{\tiny$\,\pm\,$0.01}$^{\dagger}$ & 0.572{\tiny$\,\pm\,$0.31}$^{\dagger}$ \\
        \textbf{G-2B} & $0.01$ & 0.075{\tiny$\,\pm\,$0.04}$^{\dagger}$ & 0.638{\tiny$\,\pm\,$0.11}$^{\dagger}$ & 0.097{\tiny$\,\pm\,$0.02}$^{\dagger}$ & 0.059{\tiny$\,\pm\,$0.02}$^{\dagger}$ & 0.007{\tiny$\,\pm\,$0.00} & 0.322{\tiny$\,\pm\,$0.08}$^{\dagger}$ \\
        \textbf{G-7B} & $0.01$ & 0.127{\tiny$\,\pm\,$0.10}$^{\dagger}$ & 0.624{\tiny$\,\pm\,$0.16}$^{\dagger}$ & 0.084{\tiny$\,\pm\,$0.03}$^{\dagger}$ & 0.070{\tiny$\,\pm\,$0.05}$^{\dagger}$ & 0.009{\tiny$\,\pm\,$0.01} & 0.552{\tiny$\,\pm\,$0.31}$^{\dagger}$ \\
        \textbf{G4-E4B} & $0.01$ & 0.101{\tiny$\,\pm\,$0.03}$^{\dagger}$ & 0.837{\tiny$\,\pm\,$0.10}$^{\dagger}$ & 0.178{\tiny$\,\pm\,$0.01}$^{\dagger}$ & 0.053{\tiny$\,\pm\,$0.02}$^{\dagger}$ & 0.012{\tiny$\,\pm\,$0.01}$^{**}$ & 0.384{\tiny$\,\pm\,$0.07}$^{\dagger}$ \\
        \textbf{Q-1.5B} & $0.01$ & 0.054{\tiny$\,\pm\,$0.08}$^{\dagger}$ & 0.733{\tiny$\,\pm\,$0.12}$^{\dagger}$ & 0.112{\tiny$\,\pm\,$0.02}$^{\dagger}$ & 0.033{\tiny$\,\pm\,$0.01}$^{\dagger}$ & 0.006{\tiny$\,\pm\,$0.00} & 0.343{\tiny$\,\pm\,$0.10}$^{\dagger}$ \\
        \textbf{Q-7B} & $0.01$ & 0.277{\tiny$\,\pm\,$0.14}$^{\dagger}$ & 0.673{\tiny$\,\pm\,$0.14}$^{\dagger}$ & 0.090{\tiny$\,\pm\,$0.02}$^{\dagger}$ & 0.069{\tiny$\,\pm\,$0.05}$^{\dagger}$ & 0.005{\tiny$\,\pm\,$0.00} & 0.588{\tiny$\,\pm\,$0.31}$^{\dagger}$ \\
        \textbf{Q3.5-9B} & $0.01$ & 0.063{\tiny$\,\pm\,$0.02}$^{\dagger}$ & 0.520{\tiny$\,\pm\,$0.07}$^{\dagger}$ & 0.080{\tiny$\,\pm\,$0.01}$^{\dagger}$ & 0.088{\tiny$\,\pm\,$0.03}$^{\dagger}$ & 0.006{\tiny$\,\pm\,$0.00} & 0.441{\tiny$\,\pm\,$0.06}$^{\dagger}$ \\
        \textbf{GPT-5.4} & $0.01$ & \xmark{} & 0.414{\tiny$\,\pm\,$0.13}$^{\dagger}$ & 0.056{\tiny$\,\pm\,$0.02}$^{\dagger}$ & \xmark{} & 0.154{\tiny$\,\pm\,$0.10}$^{\dagger}$ & 0.020{\tiny$\,\pm\,$0.08} \\
        \textbf{Gem-2.5F} & $0.01$ & \xmark{} & 0.312{\tiny$\,\pm\,$0.28}$^{\dagger}$ & 0.105{\tiny$\,\pm\,$0.02}$^{\dagger}$ & \xmark{} & \xmark{} & 0.005{\tiny$\,\pm\,$0.03} \\
        \textbf{Cl-S-4.6} & $0.01$ & \xmark{} & 0.505{\tiny$\,\pm\,$0.13}$^{\dagger}$ & 0.120{\tiny$\,\pm\,$0.09}$^{\dagger}$ & \xmark{} & \xmark{} & 0.061{\tiny$\,\pm\,$0.11}$^{\dagger}$ \\
        \midrule
        \multicolumn{8}{l}{\textit{\textbf{$\varepsilon=0.05$ (primary)}}} \\
        \midrule
        \textbf{L-3B} & $0.05$ & 0.124{\tiny$\,\pm\,$0.06}$^{\dagger}$ & 0.608{\tiny$\,\pm\,$0.10}$^{\dagger}$ & 0.087{\tiny$\,\pm\,$0.02}$^{\dagger}$ & 0.039{\tiny$\,\pm\,$0.01} & 0.010{\tiny$\,\pm\,$0.01} & 0.350{\tiny$\,\pm\,$0.07}$^{\dagger}$ \\
        \textbf{L-8B} & $0.05$ & 0.267{\tiny$\,\pm\,$0.12}$^{\dagger}$ & 0.630{\tiny$\,\pm\,$0.16}$^{\dagger}$ & 0.085{\tiny$\,\pm\,$0.02}$^{\dagger}$ & 0.042{\tiny$\,\pm\,$0.03} & 0.012{\tiny$\,\pm\,$0.01} & 0.572{\tiny$\,\pm\,$0.31}$^{\dagger}$ \\
        \textbf{G-2B} & $0.05$ & 0.075{\tiny$\,\pm\,$0.04}$^{\dagger}$ & 0.638{\tiny$\,\pm\,$0.11}$^{\dagger}$ & 0.097{\tiny$\,\pm\,$0.02}$^{\dagger}$ & 0.059{\tiny$\,\pm\,$0.02}$^{\dagger}$ & 0.007{\tiny$\,\pm\,$0.00} & 0.322{\tiny$\,\pm\,$0.08}$^{\dagger}$ \\
        \textbf{G-7B} & $0.05$ & 0.127{\tiny$\,\pm\,$0.10}$^{\dagger}$ & 0.624{\tiny$\,\pm\,$0.16}$^{\dagger}$ & 0.084{\tiny$\,\pm\,$0.03}$^{\dagger}$ & 0.070{\tiny$\,\pm\,$0.05}$^{\dagger}$ & 0.009{\tiny$\,\pm\,$0.01} & 0.552{\tiny$\,\pm\,$0.31}$^{\dagger}$ \\
        \textbf{G4-E4B} & $0.05$ & 0.101{\tiny$\,\pm\,$0.03}$^{\dagger}$ & 0.837{\tiny$\,\pm\,$0.10}$^{\dagger}$ & 0.178{\tiny$\,\pm\,$0.01}$^{\dagger}$ & 0.053{\tiny$\,\pm\,$0.02} & 0.012{\tiny$\,\pm\,$0.01} & 0.384{\tiny$\,\pm\,$0.07}$^{\dagger}$ \\
        \textbf{Q-1.5B} & $0.05$ & 0.054{\tiny$\,\pm\,$0.08} & 0.733{\tiny$\,\pm\,$0.12}$^{\dagger}$ & 0.112{\tiny$\,\pm\,$0.02}$^{\dagger}$ & 0.033{\tiny$\,\pm\,$0.01} & 0.006{\tiny$\,\pm\,$0.00} & 0.343{\tiny$\,\pm\,$0.10}$^{\dagger}$ \\
        \textbf{Q-7B} & $0.05$ & 0.277{\tiny$\,\pm\,$0.14}$^{\dagger}$ & 0.673{\tiny$\,\pm\,$0.14}$^{\dagger}$ & 0.090{\tiny$\,\pm\,$0.02}$^{\dagger}$ & 0.069{\tiny$\,\pm\,$0.05}$^{\dagger}$ & 0.005{\tiny$\,\pm\,$0.00} & 0.588{\tiny$\,\pm\,$0.31}$^{\dagger}$ \\
        \textbf{Q3.5-9B} & $0.05$ & 0.063{\tiny$\,\pm\,$0.02}$^{\dagger}$ & 0.520{\tiny$\,\pm\,$0.07}$^{\dagger}$ & 0.080{\tiny$\,\pm\,$0.01}$^{\dagger}$ & 0.088{\tiny$\,\pm\,$0.03}$^{\dagger}$ & 0.006{\tiny$\,\pm\,$0.00} & 0.441{\tiny$\,\pm\,$0.06}$^{\dagger}$ \\
        \textbf{GPT-5.4} & $0.05$ & \xmark{} & 0.414{\tiny$\,\pm\,$0.13}$^{\dagger}$ & 0.056{\tiny$\,\pm\,$0.02}$^{\dagger}$ & \xmark{} & 0.154{\tiny$\,\pm\,$0.10}$^{\dagger}$ & 0.020{\tiny$\,\pm\,$0.08} \\
        \textbf{Gem-2.5F} & $0.05$ & \xmark{} & 0.312{\tiny$\,\pm\,$0.28}$^{\dagger}$ & 0.105{\tiny$\,\pm\,$0.02}$^{\dagger}$ & \xmark{} & \xmark{} & 0.005{\tiny$\,\pm\,$0.03} \\
        \textbf{Cl-S-4.6} & $0.05$ & \xmark{} & 0.505{\tiny$\,\pm\,$0.13}$^{\dagger}$ & 0.120{\tiny$\,\pm\,$0.09}$^{\dagger}$ & \xmark{} & \xmark{} & 0.061{\tiny$\,\pm\,$0.11} \\
        \midrule
        \multicolumn{8}{l}{\textit{$\varepsilon=0.10$}} \\
        \midrule
        \textbf{L-3B} & $0.10$ & 0.124{\tiny$\,\pm\,$0.06}$^{\dagger}$ & 0.608{\tiny$\,\pm\,$0.10}$^{\dagger}$ & 0.087{\tiny$\,\pm\,$0.02} & 0.039{\tiny$\,\pm\,$0.01} & 0.010{\tiny$\,\pm\,$0.01} & 0.350{\tiny$\,\pm\,$0.07}$^{\dagger}$ \\
        \textbf{L-8B} & $0.10$ & 0.267{\tiny$\,\pm\,$0.12}$^{\dagger}$ & 0.630{\tiny$\,\pm\,$0.16}$^{\dagger}$ & 0.085{\tiny$\,\pm\,$0.02} & 0.042{\tiny$\,\pm\,$0.03} & 0.012{\tiny$\,\pm\,$0.01} & 0.572{\tiny$\,\pm\,$0.31}$^{\dagger}$ \\
        \textbf{G-2B} & $0.10$ & 0.075{\tiny$\,\pm\,$0.04} & 0.638{\tiny$\,\pm\,$0.11}$^{\dagger}$ & 0.097{\tiny$\,\pm\,$0.02} & 0.059{\tiny$\,\pm\,$0.02} & 0.007{\tiny$\,\pm\,$0.00} & 0.322{\tiny$\,\pm\,$0.08}$^{\dagger}$ \\
        \textbf{G-7B} & $0.10$ & 0.127{\tiny$\,\pm\,$0.10}$^{\dagger}$ & 0.624{\tiny$\,\pm\,$0.16}$^{\dagger}$ & 0.084{\tiny$\,\pm\,$0.03} & 0.070{\tiny$\,\pm\,$0.05} & 0.009{\tiny$\,\pm\,$0.01} & 0.552{\tiny$\,\pm\,$0.31}$^{\dagger}$ \\
        \textbf{G4-E4B} & $0.10$ & 0.101{\tiny$\,\pm\,$0.03} & 0.837{\tiny$\,\pm\,$0.10}$^{\dagger}$ & 0.178{\tiny$\,\pm\,$0.01}$^{\dagger}$ & 0.053{\tiny$\,\pm\,$0.02} & 0.012{\tiny$\,\pm\,$0.01} & 0.384{\tiny$\,\pm\,$0.07}$^{\dagger}$ \\
        \textbf{Q-1.5B} & $0.10$ & 0.054{\tiny$\,\pm\,$0.08} & 0.733{\tiny$\,\pm\,$0.12}$^{\dagger}$ & 0.112{\tiny$\,\pm\,$0.02}$^{\dagger}$ & 0.033{\tiny$\,\pm\,$0.01} & 0.006{\tiny$\,\pm\,$0.00} & 0.343{\tiny$\,\pm\,$0.10}$^{\dagger}$ \\
        \textbf{Q-7B} & $0.10$ & 0.277{\tiny$\,\pm\,$0.14}$^{\dagger}$ & 0.673{\tiny$\,\pm\,$0.14}$^{\dagger}$ & 0.090{\tiny$\,\pm\,$0.02} & 0.069{\tiny$\,\pm\,$0.05} & 0.005{\tiny$\,\pm\,$0.00} & 0.588{\tiny$\,\pm\,$0.31}$^{\dagger}$ \\
        \textbf{Q3.5-9B} & $0.10$ & 0.063{\tiny$\,\pm\,$0.02} & 0.520{\tiny$\,\pm\,$0.07}$^{\dagger}$ & 0.080{\tiny$\,\pm\,$0.01} & 0.088{\tiny$\,\pm\,$0.03} & 0.006{\tiny$\,\pm\,$0.00} & 0.441{\tiny$\,\pm\,$0.06}$^{\dagger}$ \\
        \textbf{GPT-5.4} & $0.10$ & \xmark{} & 0.414{\tiny$\,\pm\,$0.13}$^{\dagger}$ & 0.056{\tiny$\,\pm\,$0.02} & \xmark{} & 0.154{\tiny$\,\pm\,$0.10}$^{\dagger}$ & 0.020{\tiny$\,\pm\,$0.08} \\
        \textbf{Gem-2.5F} & $0.10$ & \xmark{} & 0.312{\tiny$\,\pm\,$0.28}$^{\dagger}$ & 0.105{\tiny$\,\pm\,$0.02}$^{*}$ & \xmark{} & \xmark{} & 0.005{\tiny$\,\pm\,$0.03} \\
        \textbf{Cl-S-4.6} & $0.10$ & \xmark{} & 0.505{\tiny$\,\pm\,$0.13}$^{\dagger}$ & 0.120{\tiny$\,\pm\,$0.09}$^{*}$ & \xmark{} & \xmark{} & 0.061{\tiny$\,\pm\,$0.11} \\
        \midrule
        \multicolumn{8}{l}{\textit{Significance counts (across all models with axis available)}} \\
        \midrule
        \multicolumn{2}{r}{\textit{$n$ available}} & 8 & 11 & 11 & 8 & 9 & 11 \\
        $\varepsilon=0.01$ & $p<.05$ & 8 & 11 & 11 & 8 & 3 & 9 \\
         & $p<.01$ & 8 & 11 & 11 & 8 & 3 & 9 \\
         & $p<.001$ & 8 & 11 & 11 & 8 & 2 & 9 \\
        \rowcolor[gray]{.85} \textbf{$\varepsilon=0.05$} & $p<.05$ & 7 & 11 & 11 & 4 & 1 & 8 \\
        \rowcolor[gray]{.85}  & $p<.01$ & 7 & 11 & 11 & 4 & 1 & 8 \\
        \rowcolor[gray]{.85}  & $p<.001$ & 7 & 11 & 11 & 4 & 1 & 8 \\
        $\varepsilon=0.10$ & $p<.05$ & 4 & 11 & 4 & 0 & 1 & 8 \\
         & $p<.01$ & 4 & 11 & 2 & 0 & 1 & 8 \\
         & $p<.001$ & 4 & 11 & 2 & 0 & 1 & 8 \\
        \bottomrule
    \end{tabular}
    
\end{table*}

\section{Full Results}
\label{app:full_results}

This appendix reports the full per-dataset
$\mathrm{SGS}_X$ results across all axes. Table~\ref{tab:sgs_open_per_dataset}
covers the eight open-source models on the six evaluation datasets and
all six axes, while Table~\ref{tab:sgs_gpt_confidence} reports the
confidence axis ($\Delta$-Ent) for GPT-5.4, the only closed-source
model that exposes the per-token probabilities required for this
metric. The full results are consistent with the patterns described
in Section~\ref{sec:exp-results}.

\begin{table}[H]
    \centering
    \caption{Per-dataset SGS for GPT-5.4 on the confidence axis $\Delta$-Ent (entropy) at $\varepsilon=0.05$. GPT-5.4 is the only closed-source model that returns the per-token probabilities required for this axis. $^{*}p<0.05$, $^{**}p<0.01$, $^{\dagger}p<0.001$ ($H_1\!:\mathrm{SGS}>\varepsilon$).}
    \label{tab:sgs_gpt_confidence}
    \scriptsize
    \setlength{\tabcolsep}{3pt}
    \renewcommand{\arraystretch}{1.05}
    \resizebox{\columnwidth}{!}{%
    \begin{tabular}{lccccccc}
        \toprule
        & TruthfulQA & NQ & Alpaca & SimpleQA-V & TriviaQA & HotpotQA & \textbf{Total} \\
        \midrule
        $\Delta$-Ent & 0.218$^{\dagger}$ & 0.184$^{\dagger}$ & 0.143$^{\dagger}$ & 0.083$^{**}$ & 0.110$^{**}$ & 0.189$^{\dagger}$ & \textbf{0.154}$^{\dagger}$ \\
        $\pm$\,s.d.  & 0.07 & 0.09 & 0.05 & 0.05 & 0.07 & 0.15 & \textbf{0.10} \\
        \bottomrule
    \end{tabular}}
\end{table}

\begin{table*}[t]
    \centering
    \tiny
     \caption{Per-dataset SGS for the eight open-source models on all six evaluation datasets, at $\varepsilon=0.05$. The shaded ``Total SGS'' row pools prompts across all datasets. $^{*}p<0.05$, $^{**}p<0.01$, $^{\dagger}p<0.001$ (one-sample one-sided $t$-test, $H_1\!:\mathrm{SGS}>\varepsilon$). \xmark{} = axis unavailable for that model.}
    \label{tab:sgs_open_per_dataset}
    \setlength{\tabcolsep}{3pt}
    \renewcommand{\arraystretch}{1.1}
    \begin{tabular}{ll cccccc}
        \toprule
        \textbf{Model} & \textbf{Dataset} & \textbf{Activation} & \multicolumn{2}{c}{\textbf{Generation Consistency}} & \multicolumn{2}{c}{\textbf{Confidence}} & \textbf{Mirroring} \\
        \cmidrule(lr){3-3} \cmidrule(lr){4-5} \cmidrule(lr){6-7} \cmidrule(lr){8-8}
         &  & $\Delta$-Cos & $\Delta$-BLEU & $\Delta$-BERT & $\Delta$-Prob & $\Delta$-Ent & $\Delta$-MR \\
        \midrule
        \multirow{6}{*}{\textbf{L-3B}}
        & TruthfulQA & 0.251{\tiny$\,\pm\,$0.05}$^{\dagger}$ & 0.672{\tiny$\,\pm\,$0.08}$^{\dagger}$ & 0.097{\tiny$\,\pm\,$0.01}$^{\dagger}$ & 0.044{\tiny$\,\pm\,$0.01} & 0.007{\tiny$\,\pm\,$0.00} & 0.309{\tiny$\,\pm\,$0.06}$^{\dagger}$ \\
        & Natural Questions & 0.108{\tiny$\,\pm\,$0.02}$^{\dagger}$ & 0.601{\tiny$\,\pm\,$0.13}$^{\dagger}$ & 0.085{\tiny$\,\pm\,$0.02}$^{\dagger}$ & 0.030{\tiny$\,\pm\,$0.01} & 0.006{\tiny$\,\pm\,$0.00} & 0.309{\tiny$\,\pm\,$0.06}$^{\dagger}$ \\
        & Alpaca & 0.092{\tiny$\,\pm\,$0.02}$^{\dagger}$ & 0.553{\tiny$\,\pm\,$0.11}$^{\dagger}$ & 0.085{\tiny$\,\pm\,$0.02}$^{\dagger}$ & 0.050{\tiny$\,\pm\,$0.02} & 0.010{\tiny$\,\pm\,$0.01} & 0.393{\tiny$\,\pm\,$0.05}$^{\dagger}$ \\
        & SimpleQA Verified & 0.106{\tiny$\,\pm\,$0.02}$^{\dagger}$ & 0.578{\tiny$\,\pm\,$0.08}$^{\dagger}$ & 0.078{\tiny$\,\pm\,$0.02}$^{\dagger}$ & 0.035{\tiny$\,\pm\,$0.01} & 0.015{\tiny$\,\pm\,$0.01} & 0.342{\tiny$\,\pm\,$0.06}$^{\dagger}$ \\
        & TriviaQA & 0.085{\tiny$\,\pm\,$0.02}$^{\dagger}$ & 0.608{\tiny$\,\pm\,$0.08}$^{\dagger}$ & 0.090{\tiny$\,\pm\,$0.01}$^{\dagger}$ & 0.039{\tiny$\,\pm\,$0.01} & 0.012{\tiny$\,\pm\,$0.00} & 0.368{\tiny$\,\pm\,$0.07}$^{\dagger}$ \\
        & HotpotQA & 0.103{\tiny$\,\pm\,$0.02}$^{\dagger}$ & 0.633{\tiny$\,\pm\,$0.05}$^{\dagger}$ & 0.088{\tiny$\,\pm\,$0.01}$^{\dagger}$ & 0.037{\tiny$\,\pm\,$0.01} & 0.013{\tiny$\,\pm\,$0.01} & 0.378{\tiny$\,\pm\,$0.08}$^{\dagger}$ \\
        \rowcolor[gray]{.85} \multicolumn{2}{r}{\textbf{Total SGS}} & \textbf{0.124{\tiny$\,\pm\,$0.06}$^{\dagger}$} & \textbf{0.608{\tiny$\,\pm\,$0.10}$^{\dagger}$} & \textbf{0.087{\tiny$\,\pm\,$0.02}$^{\dagger}$} & \textbf{0.039{\tiny$\,\pm\,$0.01}} & \textbf{0.010{\tiny$\,\pm\,$0.01}} & \textbf{0.350{\tiny$\,\pm\,$0.07}$^{\dagger}$} \\
        \midrule
        \multirow{6}{*}{\textbf{L-8B}}
        & TruthfulQA & 0.230{\tiny$\,\pm\,$0.08}$^{\dagger}$ & 0.606{\tiny$\,\pm\,$0.06}$^{\dagger}$ & 0.081{\tiny$\,\pm\,$0.01}$^{\dagger}$ & 0.030{\tiny$\,\pm\,$0.01} & 0.007{\tiny$\,\pm\,$0.00} & 0.278{\tiny$\,\pm\,$0.05}$^{\dagger}$ \\
        & Natural Questions & 0.314{\tiny$\,\pm\,$0.09}$^{\dagger}$ & 0.648{\tiny$\,\pm\,$0.16}$^{\dagger}$ & 0.085{\tiny$\,\pm\,$0.03}$^{\dagger}$ & 0.044{\tiny$\,\pm\,$0.04} & 0.013{\tiny$\,\pm\,$0.01} & 0.638{\tiny$\,\pm\,$0.33}$^{\dagger}$ \\
        & Alpaca & 0.093{\tiny$\,\pm\,$0.02}$^{\dagger}$ & 0.513{\tiny$\,\pm\,$0.13}$^{\dagger}$ & 0.078{\tiny$\,\pm\,$0.02}$^{\dagger}$ & 0.052{\tiny$\,\pm\,$0.01} & 0.009{\tiny$\,\pm\,$0.01} & 0.399{\tiny$\,\pm\,$0.07}$^{\dagger}$ \\
        & SimpleQA Verified & 0.093{\tiny$\,\pm\,$0.02}$^{\dagger}$ & 0.568{\tiny$\,\pm\,$0.10}$^{\dagger}$ & 0.077{\tiny$\,\pm\,$0.02}$^{\dagger}$ & 0.026{\tiny$\,\pm\,$0.01} & 0.010{\tiny$\,\pm\,$0.00} & 0.362{\tiny$\,\pm\,$0.08}$^{\dagger}$ \\
        & TriviaQA & 0.085{\tiny$\,\pm\,$0.02}$^{\dagger}$ & 0.546{\tiny$\,\pm\,$0.14}$^{\dagger}$ & 0.085{\tiny$\,\pm\,$0.02}$^{\dagger}$ & 0.030{\tiny$\,\pm\,$0.01} & 0.012{\tiny$\,\pm\,$0.01} & 0.395{\tiny$\,\pm\,$0.07}$^{\dagger}$ \\
        & HotpotQA & 0.084{\tiny$\,\pm\,$0.02}$^{\dagger}$ & 0.626{\tiny$\,\pm\,$0.10}$^{\dagger}$ & 0.089{\tiny$\,\pm\,$0.01}$^{\dagger}$ & 0.032{\tiny$\,\pm\,$0.01} & 0.014{\tiny$\,\pm\,$0.01} & 0.375{\tiny$\,\pm\,$0.07}$^{\dagger}$ \\
        \rowcolor[gray]{.85} \multicolumn{2}{r}{\textbf{Total SGS}} & \textbf{0.267{\tiny$\,\pm\,$0.12}$^{\dagger}$} & \textbf{0.630{\tiny$\,\pm\,$0.16}$^{\dagger}$} & \textbf{0.085{\tiny$\,\pm\,$0.02}$^{\dagger}$} & \textbf{0.042{\tiny$\,\pm\,$0.03}} & \textbf{0.012{\tiny$\,\pm\,$0.01}} & \textbf{0.572{\tiny$\,\pm\,$0.31}$^{\dagger}$} \\
        \midrule
        \multirow{6}{*}{\textbf{G-2B}}
        & TruthfulQA & 0.153{\tiny$\,\pm\,$0.05}$^{\dagger}$ & 0.737{\tiny$\,\pm\,$0.06}$^{\dagger}$ & 0.111{\tiny$\,\pm\,$0.02}$^{\dagger}$ & 0.057{\tiny$\,\pm\,$0.02} & 0.008{\tiny$\,\pm\,$0.00} & 0.248{\tiny$\,\pm\,$0.03}$^{\dagger}$ \\
        & Natural Questions & 0.079{\tiny$\,\pm\,$0.03}$^{\dagger}$ & 0.710{\tiny$\,\pm\,$0.09}$^{\dagger}$ & 0.107{\tiny$\,\pm\,$0.02}$^{\dagger}$ & 0.055{\tiny$\,\pm\,$0.02} & 0.006{\tiny$\,\pm\,$0.00} & 0.257{\tiny$\,\pm\,$0.04}$^{\dagger}$ \\
        & Alpaca & 0.051{\tiny$\,\pm\,$0.01} & 0.570{\tiny$\,\pm\,$0.08}$^{\dagger}$ & 0.094{\tiny$\,\pm\,$0.02}$^{\dagger}$ & 0.075{\tiny$\,\pm\,$0.03}$^{**}$ & 0.006{\tiny$\,\pm\,$0.00} & 0.389{\tiny$\,\pm\,$0.09}$^{\dagger}$ \\
        & SimpleQA Verified & 0.044{\tiny$\,\pm\,$0.01} & 0.569{\tiny$\,\pm\,$0.11}$^{\dagger}$ & 0.079{\tiny$\,\pm\,$0.02}$^{\dagger}$ & 0.045{\tiny$\,\pm\,$0.02} & 0.006{\tiny$\,\pm\,$0.00} & 0.333{\tiny$\,\pm\,$0.08}$^{\dagger}$ \\
        & TriviaQA & 0.068{\tiny$\,\pm\,$0.01}$^{\dagger}$ & 0.602{\tiny$\,\pm\,$0.10}$^{\dagger}$ & 0.096{\tiny$\,\pm\,$0.02}$^{\dagger}$ & 0.062{\tiny$\,\pm\,$0.02}$^{*}$ & 0.007{\tiny$\,\pm\,$0.00} & 0.368{\tiny$\,\pm\,$0.06}$^{\dagger}$ \\
        & HotpotQA & 0.055{\tiny$\,\pm\,$0.01} & 0.641{\tiny$\,\pm\,$0.10}$^{\dagger}$ & 0.097{\tiny$\,\pm\,$0.02}$^{\dagger}$ & 0.060{\tiny$\,\pm\,$0.02} & 0.009{\tiny$\,\pm\,$0.00} & 0.338{\tiny$\,\pm\,$0.07}$^{\dagger}$ \\
        \rowcolor[gray]{.85} \multicolumn{2}{r}{\textbf{Total SGS}} & \textbf{0.075{\tiny$\,\pm\,$0.04}$^{\dagger}$} & \textbf{0.638{\tiny$\,\pm\,$0.11}$^{\dagger}$} & \textbf{0.097{\tiny$\,\pm\,$0.02}$^{\dagger}$} & \textbf{0.059{\tiny$\,\pm\,$0.02}$^{\dagger}$} & \textbf{0.007{\tiny$\,\pm\,$0.00}} & \textbf{0.322{\tiny$\,\pm\,$0.08}$^{\dagger}$} \\
        \midrule
        \multirow{6}{*}{\textbf{G-7B}}
        & TruthfulQA & 0.100{\tiny$\,\pm\,$0.06}$^{**}$ & 0.646{\tiny$\,\pm\,$0.12}$^{\dagger}$ & 0.091{\tiny$\,\pm\,$0.02}$^{\dagger}$ & 0.057{\tiny$\,\pm\,$0.02} & 0.014{\tiny$\,\pm\,$0.01} & 0.233{\tiny$\,\pm\,$0.04}$^{\dagger}$ \\
        & Natural Questions & 0.153{\tiny$\,\pm\,$0.11}$^{\dagger}$ & 0.637{\tiny$\,\pm\,$0.18}$^{\dagger}$ & 0.082{\tiny$\,\pm\,$0.03}$^{\dagger}$ & 0.071{\tiny$\,\pm\,$0.06}$^{\dagger}$ & 0.009{\tiny$\,\pm\,$0.01} & 0.616{\tiny$\,\pm\,$0.33}$^{\dagger}$ \\
        & Alpaca & 0.011{\tiny$\,\pm\,$0.01} & 0.553{\tiny$\,\pm\,$0.11}$^{\dagger}$ & 0.091{\tiny$\,\pm\,$0.01}$^{\dagger}$ & 0.084{\tiny$\,\pm\,$0.04}$^{**}$ & 0.005{\tiny$\,\pm\,$0.00} & 0.398{\tiny$\,\pm\,$0.08}$^{\dagger}$ \\
        & SimpleQA Verified & 0.029{\tiny$\,\pm\,$0.02} & 0.548{\tiny$\,\pm\,$0.11}$^{\dagger}$ & 0.081{\tiny$\,\pm\,$0.02}$^{\dagger}$ & 0.070{\tiny$\,\pm\,$0.04}$^{*}$ & 0.008{\tiny$\,\pm\,$0.00} & 0.343{\tiny$\,\pm\,$0.10}$^{\dagger}$ \\
        & TriviaQA & 0.034{\tiny$\,\pm\,$0.02} & 0.567{\tiny$\,\pm\,$0.10}$^{\dagger}$ & 0.089{\tiny$\,\pm\,$0.01}$^{\dagger}$ & 0.049{\tiny$\,\pm\,$0.02} & 0.008{\tiny$\,\pm\,$0.00} & 0.402{\tiny$\,\pm\,$0.05}$^{\dagger}$ \\
        & HotpotQA & 0.048{\tiny$\,\pm\,$0.03} & 0.602{\tiny$\,\pm\,$0.09}$^{\dagger}$ & 0.095{\tiny$\,\pm\,$0.02}$^{\dagger}$ & 0.074{\tiny$\,\pm\,$0.03}$^{**}$ & 0.008{\tiny$\,\pm\,$0.00} & 0.358{\tiny$\,\pm\,$0.09}$^{\dagger}$ \\
        \rowcolor[gray]{.85} \multicolumn{2}{r}{\textbf{Total SGS}} & \textbf{0.127{\tiny$\,\pm\,$0.10}$^{\dagger}$} & \textbf{0.624{\tiny$\,\pm\,$0.16}$^{\dagger}$} & \textbf{0.084{\tiny$\,\pm\,$0.03}$^{\dagger}$} & \textbf{0.070{\tiny$\,\pm\,$0.05}$^{\dagger}$} & \textbf{0.009{\tiny$\,\pm\,$0.01}} & \textbf{0.552{\tiny$\,\pm\,$0.31}$^{\dagger}$} \\
        \midrule
        \multirow{6}{*}{\textbf{G4-E4B}}
        & TruthfulQA & 0.074{\tiny$\,\pm\,$0.02}$^{\dagger}$ & 0.873{\tiny$\,\pm\,$0.02}$^{\dagger}$ & 0.180{\tiny$\,\pm\,$0.01}$^{\dagger}$ & 0.053{\tiny$\,\pm\,$0.03} & 0.012{\tiny$\,\pm\,$0.01} & 0.377{\tiny$\,\pm\,$0.08}$^{\dagger}$ \\
        & Natural Questions & 0.109{\tiny$\,\pm\,$0.02}$^{\dagger}$ & 0.837{\tiny$\,\pm\,$0.22}$^{\dagger}$ & 0.184{\tiny$\,\pm\,$0.01}$^{\dagger}$ & 0.053{\tiny$\,\pm\,$0.03} & 0.020{\tiny$\,\pm\,$0.01} & 0.363{\tiny$\,\pm\,$0.07}$^{\dagger}$ \\
        & Alpaca & 0.124{\tiny$\,\pm\,$0.04}$^{\dagger}$ & 0.774{\tiny$\,\pm\,$0.05}$^{\dagger}$ & 0.169{\tiny$\,\pm\,$0.00}$^{\dagger}$ & 0.072{\tiny$\,\pm\,$0.03}$^{**}$ & 0.010{\tiny$\,\pm\,$0.01} & 0.436{\tiny$\,\pm\,$0.04}$^{\dagger}$ \\
        & SimpleQA Verified & 0.092{\tiny$\,\pm\,$0.03}$^{\dagger}$ & 0.850{\tiny$\,\pm\,$0.03}$^{\dagger}$ & 0.176{\tiny$\,\pm\,$0.00}$^{\dagger}$ & 0.038{\tiny$\,\pm\,$0.01} & 0.011{\tiny$\,\pm\,$0.01} & 0.365{\tiny$\,\pm\,$0.09}$^{\dagger}$ \\
        & TriviaQA & 0.111{\tiny$\,\pm\,$0.02}$^{\dagger}$ & 0.825{\tiny$\,\pm\,$0.03}$^{\dagger}$ & 0.176{\tiny$\,\pm\,$0.01}$^{\dagger}$ & 0.059{\tiny$\,\pm\,$0.02}$^{*}$ & 0.012{\tiny$\,\pm\,$0.01} & 0.394{\tiny$\,\pm\,$0.06}$^{\dagger}$ \\
        & HotpotQA & 0.097{\tiny$\,\pm\,$0.03}$^{\dagger}$ & 0.863{\tiny$\,\pm\,$0.03}$^{\dagger}$ & 0.180{\tiny$\,\pm\,$0.01}$^{\dagger}$ & 0.043{\tiny$\,\pm\,$0.01} & 0.008{\tiny$\,\pm\,$0.00} & 0.368{\tiny$\,\pm\,$0.06}$^{\dagger}$ \\
        \rowcolor[gray]{.85} \multicolumn{2}{r}{\textbf{Total SGS}} & \textbf{0.101{\tiny$\,\pm\,$0.03}$^{\dagger}$} & \textbf{0.837{\tiny$\,\pm\,$0.10}$^{\dagger}$} & \textbf{0.178{\tiny$\,\pm\,$0.01}$^{\dagger}$} & \textbf{0.053{\tiny$\,\pm\,$0.02}} & \textbf{0.012{\tiny$\,\pm\,$0.01}} & \textbf{0.384{\tiny$\,\pm\,$0.07}$^{\dagger}$} \\
        \midrule
        \multirow{6}{*}{\textbf{Q-1.5B}}
        & TruthfulQA & 0.214{\tiny$\,\pm\,$0.07}$^{\dagger}$ & 0.818{\tiny$\,\pm\,$0.04}$^{\dagger}$ & 0.117{\tiny$\,\pm\,$0.01}$^{\dagger}$ & 0.024{\tiny$\,\pm\,$0.01} & 0.004{\tiny$\,\pm\,$0.00} & 0.253{\tiny$\,\pm\,$0.03}$^{\dagger}$ \\
        & Natural Questions & 0.027{\tiny$\,\pm\,$0.01} & 0.800{\tiny$\,\pm\,$0.09}$^{\dagger}$ & 0.118{\tiny$\,\pm\,$0.02}$^{\dagger}$ & 0.024{\tiny$\,\pm\,$0.01} & 0.005{\tiny$\,\pm\,$0.00} & 0.243{\tiny$\,\pm\,$0.02}$^{\dagger}$ \\
        & Alpaca & 0.023{\tiny$\,\pm\,$0.01} & 0.607{\tiny$\,\pm\,$0.15}$^{\dagger}$ & 0.100{\tiny$\,\pm\,$0.02}$^{\dagger}$ & 0.048{\tiny$\,\pm\,$0.01} & 0.008{\tiny$\,\pm\,$0.00} & 0.451{\tiny$\,\pm\,$0.09}$^{\dagger}$ \\
        & SimpleQA Verified & 0.016{\tiny$\,\pm\,$0.01} & 0.735{\tiny$\,\pm\,$0.11}$^{\dagger}$ & 0.111{\tiny$\,\pm\,$0.02}$^{\dagger}$ & 0.028{\tiny$\,\pm\,$0.01} & 0.007{\tiny$\,\pm\,$0.00} & 0.374{\tiny$\,\pm\,$0.08}$^{\dagger}$ \\
        & TriviaQA & 0.024{\tiny$\,\pm\,$0.00} & 0.662{\tiny$\,\pm\,$0.10}$^{\dagger}$ & 0.106{\tiny$\,\pm\,$0.02}$^{\dagger}$ & 0.034{\tiny$\,\pm\,$0.01} & 0.008{\tiny$\,\pm\,$0.00} & 0.382{\tiny$\,\pm\,$0.07}$^{\dagger}$ \\
        & HotpotQA & 0.018{\tiny$\,\pm\,$0.00} & 0.775{\tiny$\,\pm\,$0.07}$^{\dagger}$ & 0.120{\tiny$\,\pm\,$0.01}$^{\dagger}$ & 0.037{\tiny$\,\pm\,$0.01} & 0.008{\tiny$\,\pm\,$0.01} & 0.357{\tiny$\,\pm\,$0.08}$^{\dagger}$ \\
        \rowcolor[gray]{.85} \multicolumn{2}{r}{\textbf{Total SGS}} & \textbf{0.054{\tiny$\,\pm\,$0.08}} & \textbf{0.733{\tiny$\,\pm\,$0.12}$^{\dagger}$} & \textbf{0.112{\tiny$\,\pm\,$0.02}$^{\dagger}$} & \textbf{0.033{\tiny$\,\pm\,$0.01}} & \textbf{0.006{\tiny$\,\pm\,$0.00}} & \textbf{0.343{\tiny$\,\pm\,$0.10}$^{\dagger}$} \\
        \midrule
        \multirow{6}{*}{\textbf{Q-7B}}
        & TruthfulQA & 0.229{\tiny$\,\pm\,$0.09}$^{\dagger}$ & 0.705{\tiny$\,\pm\,$0.07}$^{\dagger}$ & 0.093{\tiny$\,\pm\,$0.01}$^{\dagger}$ & 0.049{\tiny$\,\pm\,$0.02} & 0.006{\tiny$\,\pm\,$0.00} & 0.379{\tiny$\,\pm\,$0.06}$^{\dagger}$ \\
        & Natural Questions & 0.335{\tiny$\,\pm\,$0.09}$^{\dagger}$ & 0.686{\tiny$\,\pm\,$0.15}$^{\dagger}$ & 0.089{\tiny$\,\pm\,$0.02}$^{\dagger}$ & 0.069{\tiny$\,\pm\,$0.06}$^{\dagger}$ & 0.005{\tiny$\,\pm\,$0.00} & 0.640{\tiny$\,\pm\,$0.33}$^{\dagger}$ \\
        & Alpaca & 0.059{\tiny$\,\pm\,$0.01}$^{**}$ & 0.562{\tiny$\,\pm\,$0.11}$^{\dagger}$ & 0.091{\tiny$\,\pm\,$0.02}$^{\dagger}$ & 0.090{\tiny$\,\pm\,$0.03}$^{\dagger}$ & 0.005{\tiny$\,\pm\,$0.00} & 0.520{\tiny$\,\pm\,$0.07}$^{\dagger}$ \\
        & SimpleQA Verified & 0.050{\tiny$\,\pm\,$0.01} & 0.595{\tiny$\,\pm\,$0.15}$^{\dagger}$ & 0.084{\tiny$\,\pm\,$0.02}$^{\dagger}$ & 0.043{\tiny$\,\pm\,$0.01} & 0.005{\tiny$\,\pm\,$0.00} & 0.375{\tiny$\,\pm\,$0.05}$^{\dagger}$ \\
        & TriviaQA & 0.061{\tiny$\,\pm\,$0.01}$^{\dagger}$ & 0.621{\tiny$\,\pm\,$0.13}$^{\dagger}$ & 0.093{\tiny$\,\pm\,$0.02}$^{\dagger}$ & 0.077{\tiny$\,\pm\,$0.02}$^{\dagger}$ & 0.006{\tiny$\,\pm\,$0.00} & 0.436{\tiny$\,\pm\,$0.05}$^{\dagger}$ \\
        & HotpotQA & 0.059{\tiny$\,\pm\,$0.02}$^{*}$ & 0.683{\tiny$\,\pm\,$0.08}$^{\dagger}$ & 0.101{\tiny$\,\pm\,$0.01}$^{\dagger}$ & 0.075{\tiny$\,\pm\,$0.02}$^{\dagger}$ & 0.006{\tiny$\,\pm\,$0.00} & 0.403{\tiny$\,\pm\,$0.06}$^{\dagger}$ \\
        \rowcolor[gray]{.85} \multicolumn{2}{r}{\textbf{Total SGS}} & \textbf{0.277{\tiny$\,\pm\,$0.14}$^{\dagger}$} & \textbf{0.673{\tiny$\,\pm\,$0.14}$^{\dagger}$} & \textbf{0.090{\tiny$\,\pm\,$0.02}$^{\dagger}$} & \textbf{0.069{\tiny$\,\pm\,$0.05}$^{\dagger}$} & \textbf{0.005{\tiny$\,\pm\,$0.00}} & \textbf{0.588{\tiny$\,\pm\,$0.31}$^{\dagger}$} \\
        \midrule
        \multirow{6}{*}{\textbf{Q3.5-9B}}
        & TruthfulQA & 0.052{\tiny$\,\pm\,$0.01} & 0.523{\tiny$\,\pm\,$0.06}$^{\dagger}$ & 0.079{\tiny$\,\pm\,$0.01}$^{\dagger}$ & 0.103{\tiny$\,\pm\,$0.02}$^{\dagger}$ & 0.005{\tiny$\,\pm\,$0.00} & 0.447{\tiny$\,\pm\,$0.03}$^{\dagger}$ \\
        & Natural Questions & 0.050{\tiny$\,\pm\,$0.01} & 0.564{\tiny$\,\pm\,$0.07}$^{\dagger}$ & 0.084{\tiny$\,\pm\,$0.01}$^{\dagger}$ & 0.105{\tiny$\,\pm\,$0.03}$^{\dagger}$ & 0.006{\tiny$\,\pm\,$0.00} & 0.491{\tiny$\,\pm\,$0.03}$^{\dagger}$ \\
        & Alpaca & 0.046{\tiny$\,\pm\,$0.02} & 0.490{\tiny$\,\pm\,$0.08}$^{\dagger}$ & 0.082{\tiny$\,\pm\,$0.02}$^{\dagger}$ & 0.105{\tiny$\,\pm\,$0.03}$^{\dagger}$ & 0.007{\tiny$\,\pm\,$0.00} & 0.473{\tiny$\,\pm\,$0.07}$^{\dagger}$ \\
        & SimpleQA Verified & 0.085{\tiny$\,\pm\,$0.02}$^{\dagger}$ & 0.486{\tiny$\,\pm\,$0.05}$^{\dagger}$ & 0.071{\tiny$\,\pm\,$0.01}$^{\dagger}$ & 0.067{\tiny$\,\pm\,$0.01}$^{\dagger}$ & 0.004{\tiny$\,\pm\,$0.00} & 0.403{\tiny$\,\pm\,$0.06}$^{\dagger}$ \\
        & TriviaQA & 0.069{\tiny$\,\pm\,$0.02}$^{**}$ & 0.544{\tiny$\,\pm\,$0.04}$^{\dagger}$ & 0.087{\tiny$\,\pm\,$0.01}$^{\dagger}$ & 0.075{\tiny$\,\pm\,$0.02}$^{\dagger}$ & 0.006{\tiny$\,\pm\,$0.00} & 0.437{\tiny$\,\pm\,$0.05}$^{\dagger}$ \\
        & HotpotQA & 0.074{\tiny$\,\pm\,$0.02}$^{\dagger}$ & 0.513{\tiny$\,\pm\,$0.06}$^{\dagger}$ & 0.077{\tiny$\,\pm\,$0.01}$^{\dagger}$ & 0.076{\tiny$\,\pm\,$0.02}$^{\dagger}$ & 0.006{\tiny$\,\pm\,$0.00} & 0.394{\tiny$\,\pm\,$0.07}$^{\dagger}$ \\
        \rowcolor[gray]{.85} \multicolumn{2}{r}{\textbf{Total SGS}} & \textbf{0.063{\tiny$\,\pm\,$0.02}$^{\dagger}$} & \textbf{0.520{\tiny$\,\pm\,$0.07}$^{\dagger}$} & \textbf{0.080{\tiny$\,\pm\,$0.01}$^{\dagger}$} & \textbf{0.088{\tiny$\,\pm\,$0.03}$^{\dagger}$} & \textbf{0.006{\tiny$\,\pm\,$0.00}} & \textbf{0.441{\tiny$\,\pm\,$0.06}$^{\dagger}$} \\
        \bottomrule
    \end{tabular}
   
\end{table*}

\section{Prompt Sample Size Ablation}
\label{subsec:ablation}

\begin{figure*}[t]
  \centering
  \includegraphics[width=\linewidth]{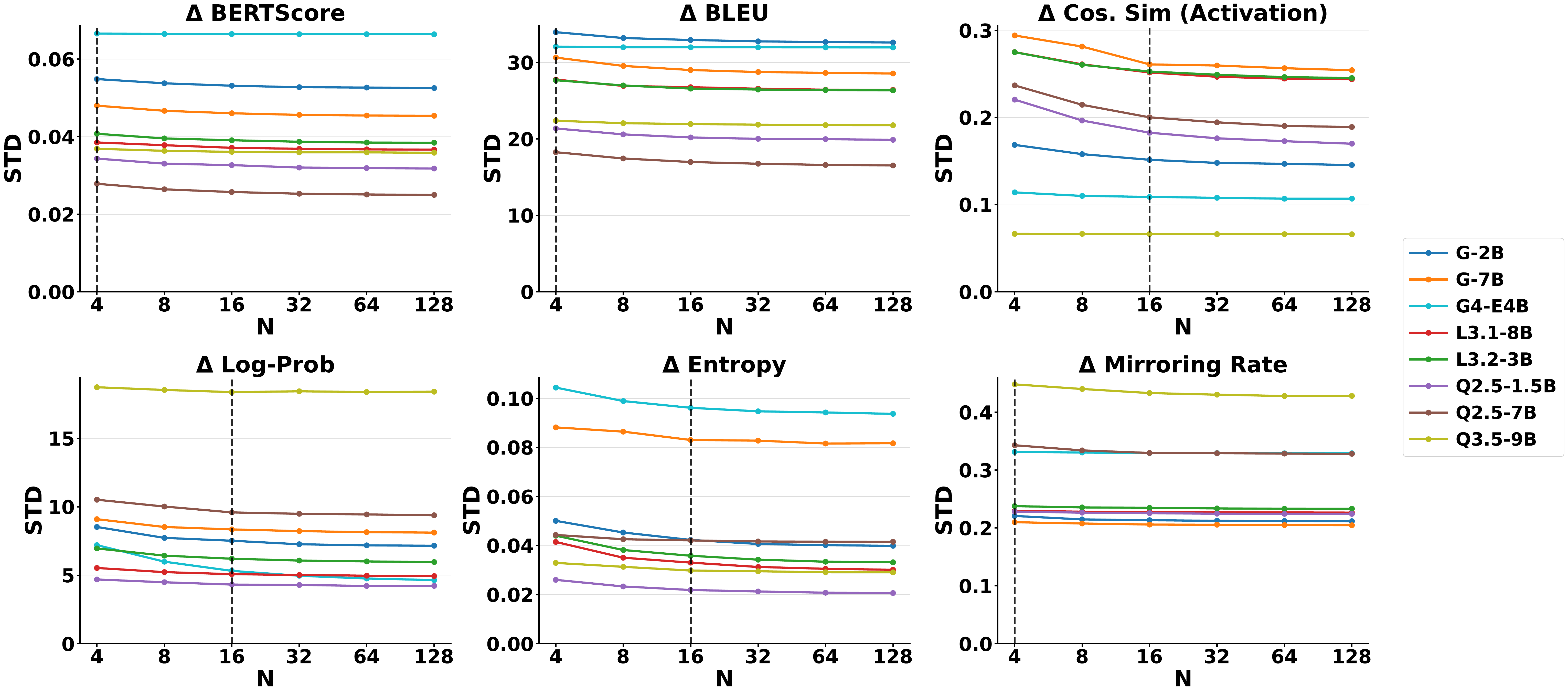}
  \caption{Per-prompt instability standard deviation as a
  function of prompt sample size $N$. Each panel shows one
  behavioral axis; each line is one model. Curves plateau by
  $N{=}16$ and remain flat through $N{=}128$, confirming that
  $N{=}16$ is sufficient for stable estimation. The dashed
  vertical line marks the selected operating point.}
  \label{fig:Nablation}
\end{figure*}

To determine the minimum number of prompts $N$ sufficient for
stable estimation of the per-prompt instability $s_X(i)$
(Eq.~\ref{eq:per_prompt}), we subsample
$N \in \{4, 8, 16, 32, 64, 128\}$ prompts without replacement
from TruthfulQA~\citep{lin2022truthfulqa}, repeating each draw
$500$ times. For each subsample we compute the standard
deviation of per-$(F, u, s, p)$ group means across the $N$
sampled prompts, and track its
bootstrap~\citep{davison1997bootstrap} distribution. We define
the recommended $N$ as the smallest value for which the
bootstrap band (P90$-$P10) stabilizes, taking the most
conservative $N$ across all models and axes.
Figure~\ref{fig:Nablation} shows that five of six axes plateau
by $N{=}16$ and remain flat through $N{=}128$, confirming that
$N{=}16$ is sufficient for stable estimation of $\mathrm{SGS}_X$.
We therefore stopped data collection at $N{=}128$ once stability
was confirmed, and adopt $N{=}16$ as our operating point.

\section{Sensitivity Analysis}
\label{app:sensitivity}

The ablations in this section confirm two design choices made in
Section~\ref{sec:exp-setting}: that all three placement positions
$p \in \mathcal{P} = \{\textsc{prefix}, \textsc{suffix},
\textsc{global}\}$ and all evaluated strength levels
$s \in \mathcal{S}_F$ induce meaningfully distinct behavioral
effects, and must therefore be retained in the variation set
$\mathcal{V}$ (Eq.~\ref{eq:variation_set}) to obtain a complete
picture of stylistic generalization. All experiments in this
section are conducted on TruthfulQA~\citep{lin2022truthfulqa}.

\subsection{Placement Ablation}
\label{app:sensitivity_place}

\begin{figure*}[t]
  \centering
  \includegraphics[width=\linewidth]{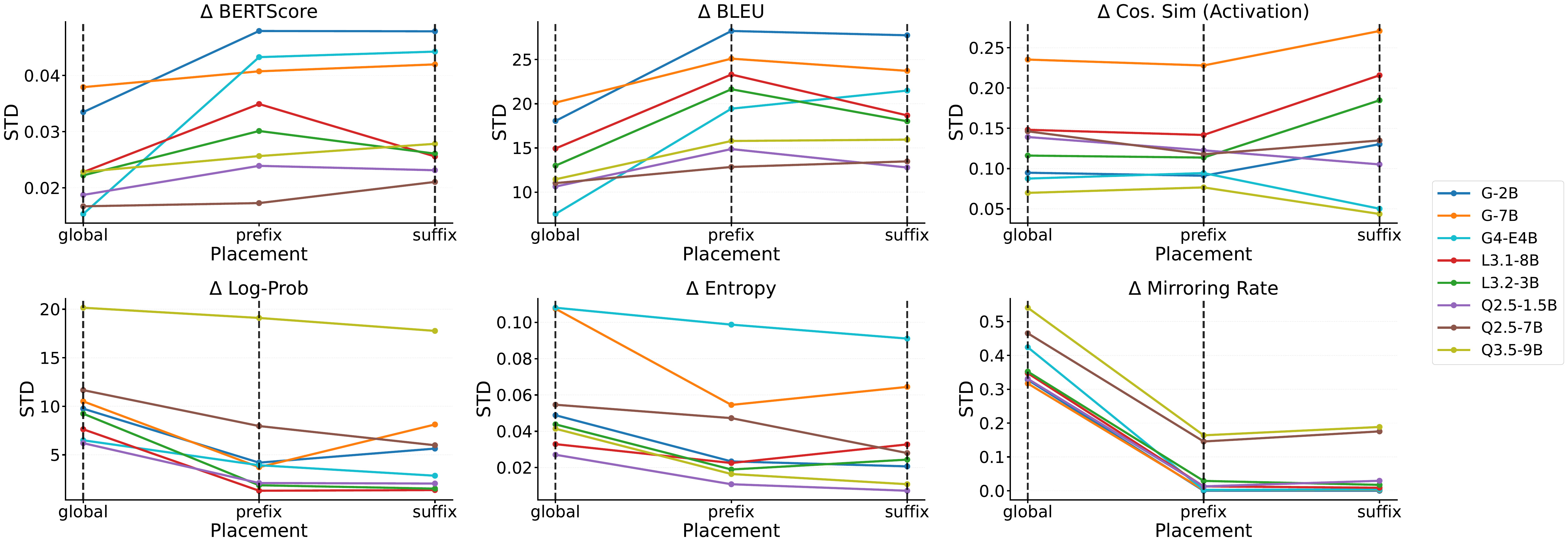}
  \caption{Per-prompt instability standard deviation by variant
  position $p$. Each panel shows one behavioral axis; each line
  is one model. The three positions
  ($\textsc{global}, \textsc{prefix}, \textsc{suffix}$) produce
  distinct profiles across axes, confirming that all placements
  contribute independent behavioral signal to $\mathrm{SGS}_X$.}
  \label{fig:sensitivity_place_std}
\end{figure*}

Figure~\ref{fig:sensitivity_place_std} shows the standard
deviation of per-$(F, u, s)$ group means as a function of
position $p$ for each axis. Across all axes and models, the
three positions produce systematically different values: global
variants produce the largest effects on generation consistency and
mirroring, while prefix and suffix placements show distinct
sensitivity profiles on the confidence and activation axes. No
single placement dominates across all axes and models, confirming
that collapsing across positions would lose behavioral signal
and that all three must be included in $\mathcal{V}$.

\subsection{Strength Ablations}
\label{app:sensitivity_strength}

\begin{figure*}[t]
  \centering
  \includegraphics[width=\linewidth]{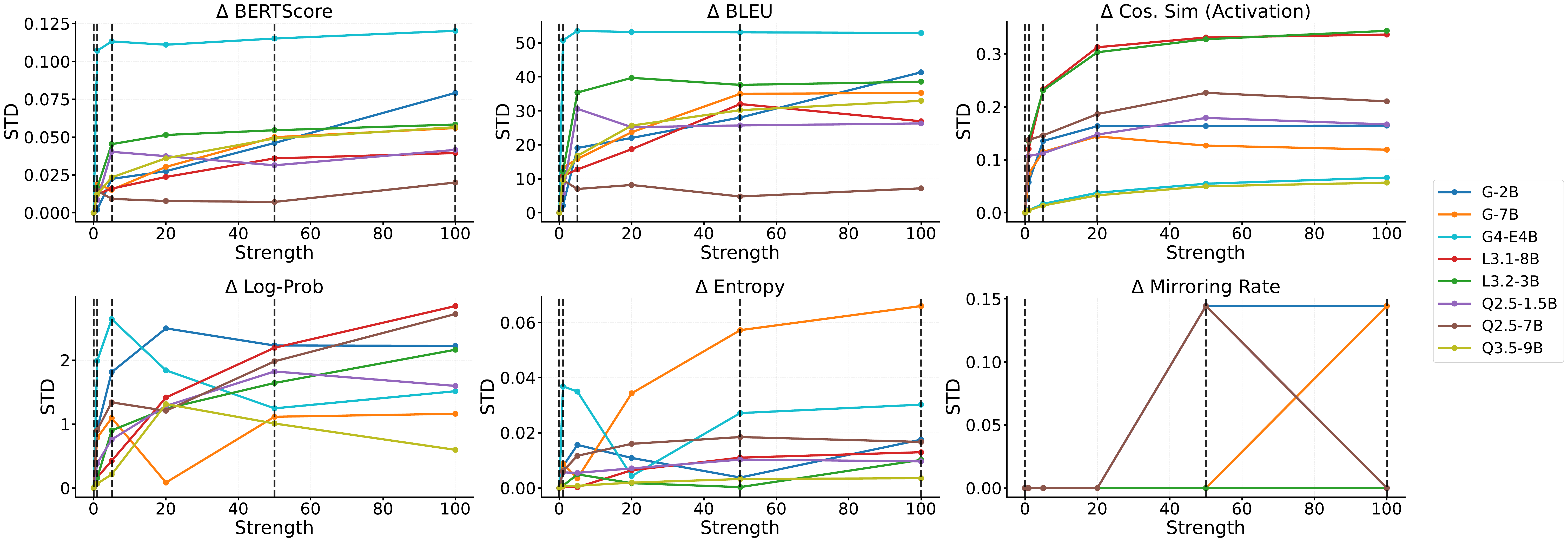}
  \caption{Per-prompt instability standard deviation vs.\ spacing
  strength (number of extra whitespace tokens injected). Effects
  grow with strength and show model-dependent saturation points,
  with generation consistency axes most responsive at moderate
  strengths.}
  \label{fig:sensitivity_strength_std_spacing}
\end{figure*}

\begin{figure*}[t]
  \centering
  \includegraphics[width=\linewidth]{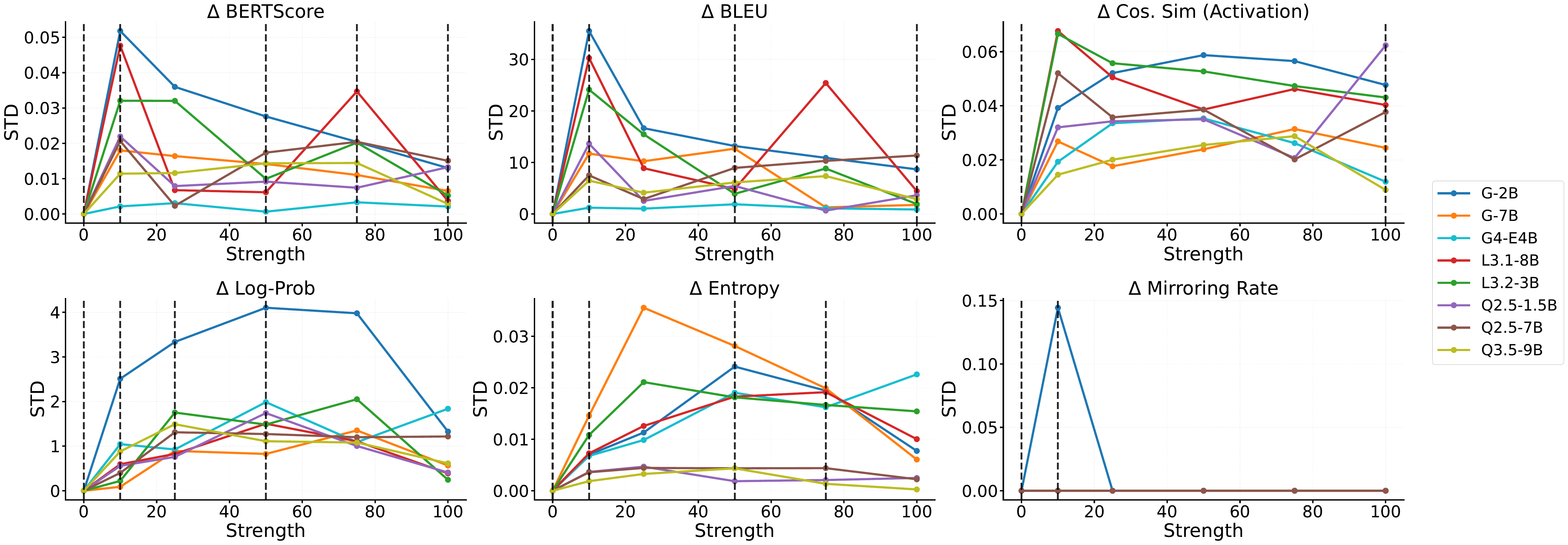}
  \caption{Per-prompt instability standard deviation vs.\
  letter-case strength (fraction of characters uppercased).
  Behavioral detectability grows with strength across most axes
  and plateaus above strength $50$, with activation geometry
  showing the largest and most consistent response.}
  \label{fig:sensitivity_strength_std_letter_case}
\end{figure*}

\begin{figure*}[t]
  \centering
  \includegraphics[width=\linewidth]{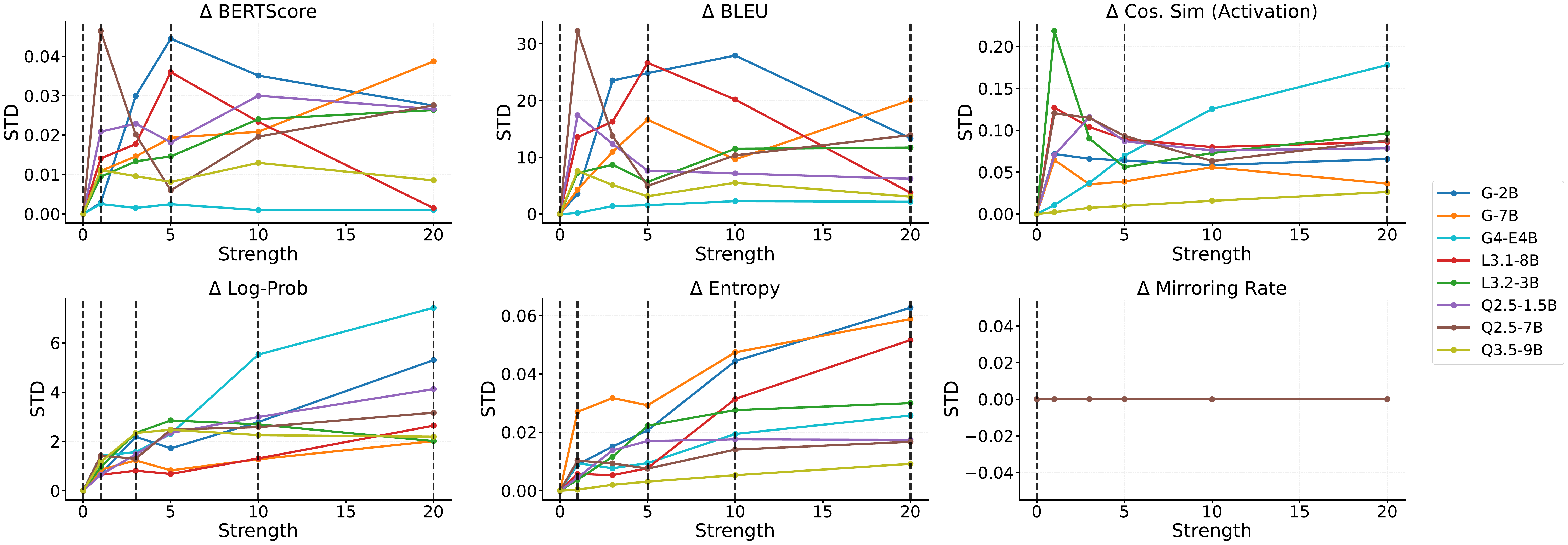}
  \caption{Per-prompt instability standard deviation vs.\
  punctuation strength (number of punctuation tokens injected
  per prompt). Generation consistency and activation axes show the
  clearest strength-dependent effects; confidence axes remain
  largely flat, consistent with their low overall
  $\mathrm{SGS}_X$ in the main results.}
  \label{fig:sensitivity_strength_std_punctuation}
\end{figure*}

\begin{figure*}[t]
  \centering
  \includegraphics[width=\linewidth]{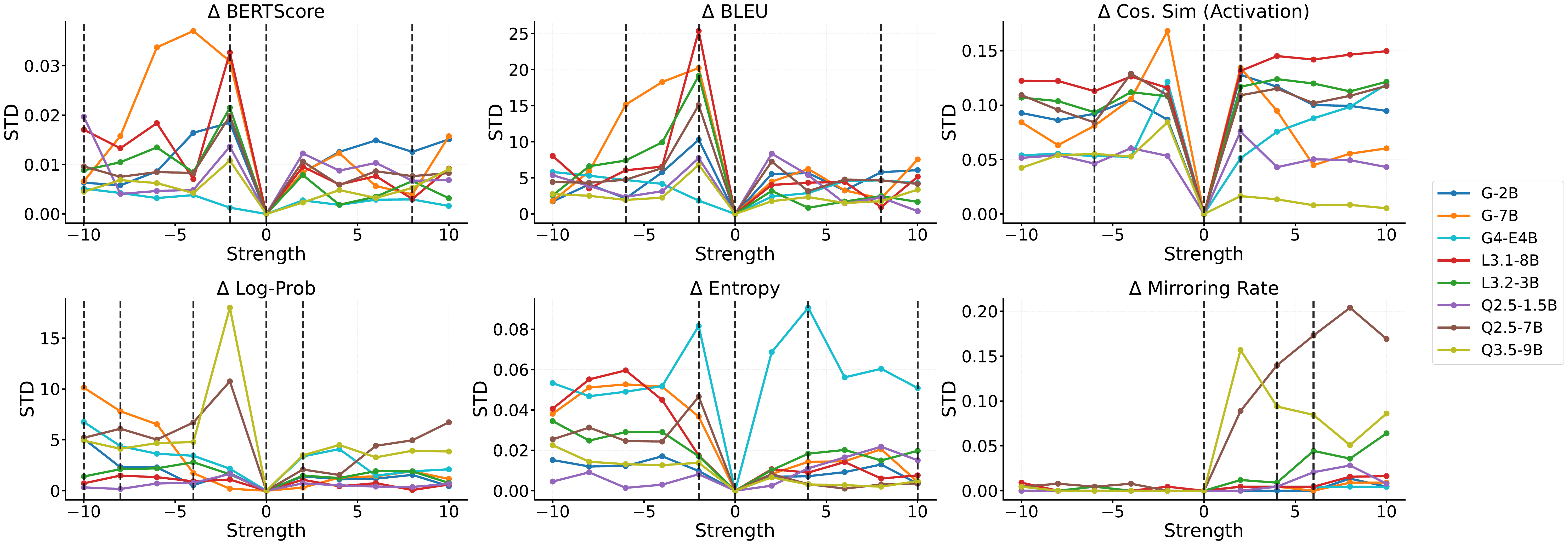}
  \caption{Per-prompt instability standard deviation vs.\
  politeness strength. Negative values correspond to rude
  framing; positive values to polite framing; zero is the
  unperturbed baseline. Rude variants produce larger and more
  consistent behavioral shifts than polite ones across most
  models and axes, indicating an asymmetry in how models
  respond to negative versus positive social tone.}
  \label{fig:sensitivity_strength_std_politeness}
\end{figure*}

\begin{figure*}[t]
  \centering
  \includegraphics[width=\linewidth]{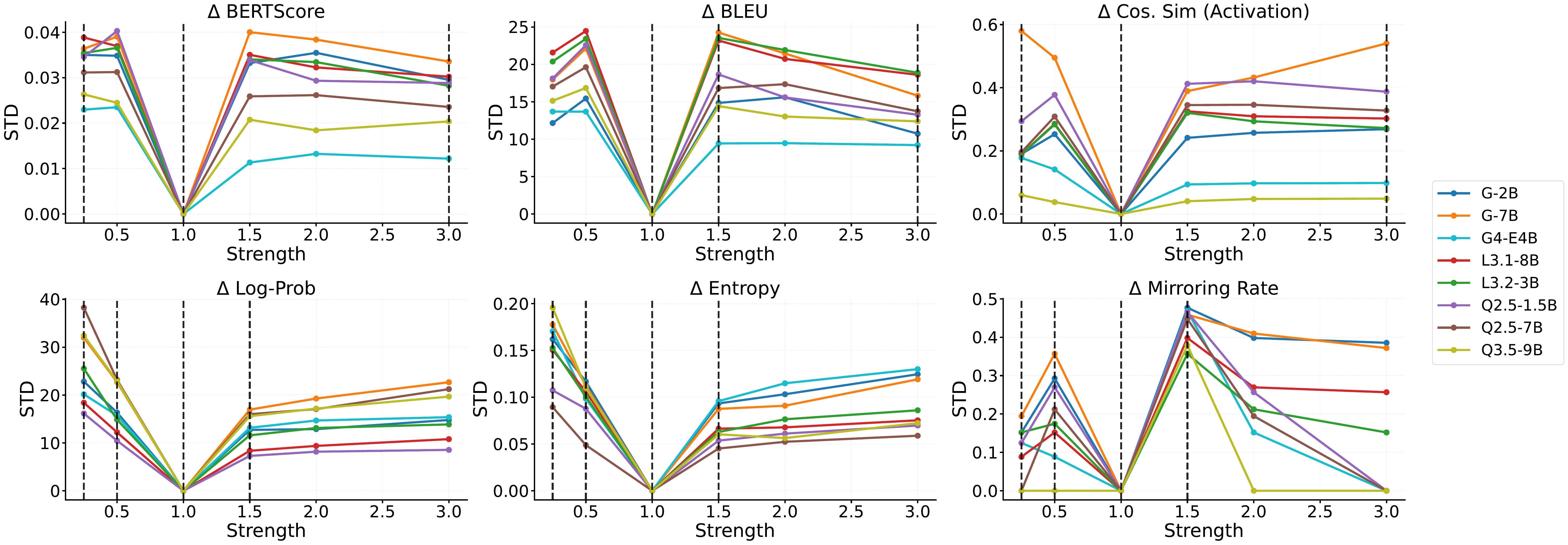}
  \caption{Per-prompt instability standard deviation vs.\
  length-variation factor (target length ratio relative to the
  original prompt; values below $1.0$ indicate compression,
  above $1.0$ indicate expansion). Compression induces larger
  and more model-consistent behavioral shifts than expansion
  across the generation consistency and mirroring axes, while
  confidence axes respond more strongly to expansion.}
  \label{fig:sensitivity_strength_std_length_variation}
\end{figure*}

Figures~\ref{fig:sensitivity_strength_std_spacing}--\ref{fig:sensitivity_strength_std_length_variation}
show the standard deviation of per-$(F, u, p)$ group means as a
function of strength $s$ for each of the five subtypes. In all
cases, different strength levels produce meaningfully different
values: effects are not monotone and do not saturate uniformly
across models and axes, indicating that strength is a genuine
source of behavioral variation rather than a simple scaling
factor. Including all evaluated strength levels in $\mathcal{V}$
is therefore necessary to capture the full range of stylistic
generalization failures.

\end{document}